\pdfoutput=1
\documentclass[11pt]{article}

\usepackage[preprint]{acl}
\usepackage{times}
\usepackage{latexsym}

\usepackage[T1]{fontenc}
\usepackage[utf8]{inputenc}

\usepackage{microtype}

\usepackage{inconsolata}

\usepackage{graphicx}
\usepackage{multirow}

\usepackage{xcolor}
\usepackage{soul}
\usepackage{fontawesome5}
\usepackage{hyperref}

\makeatletter
\newcommand{\github}[1]{%
   \href{#1}{\faGithub}%
}
\makeatother

\title{elsciRL: Integrating Language Solutions into Reinforcement Learning Problem Settings}

\author{Philip Osborne$^{1}$, Danilo S. Carvalho$^{2}$, Andr\'{e} Freitas$^{2,3,4}$\\
National Biomarker Centre, CRUK-MI, University of Manchester, United Kingdom$^{2}$\\
Department of Computer Science, University of Manchester, United Kingdom$^{3}$ \\
Idiap Research Institute, Switzerland$^{4}$ \\
\faGithub \textit{ }
    \href{https://github.com/pdfosborne/elsciRL}{elsciRL}$^{1}$ ~|~ 
\faYoutube~~\href{https://youtu.be/yPvY2RfFoRY}{Short Video}
}

\begin{document}
\maketitle
\begin{abstract}

We present elsciRL, an open-source Python library to facilitate the application of language solutions on reinforcement learning problems. We demonstrate the potential of our software by extending the Language Adapter with Self-Completing Instruction framework defined in \citep{Osborne2024} with the use of LLMs. Our approach can be re-applied to new applications with minimal setup requirements. We provide a novel GUI that allows a user to provide text input for an LLM to generate instructions which it can then self-complete. Empirical results indicate that these instructions \textit{can} improve a reinforcement learning agent's performance. Therefore, we present this work to accelerate the evaluation of language solutions on reward based environments to enable new opportunities for scientific discovery.

\end{abstract}

\section{Introduction}

Applying reinforcement learning (RL) to problems typically requires custom software for each application. For example, CleanRL \citep{huang2022cleanrl} offers single-file implementations of range of RL agent methodologies thereby separating the software for each agent and problem setting. RL agent libraries, such as Stablebaselines3 \citep{stable-baselines3}, MushroomRL \citep{MushroomRL:2021} and RLlib \citep{RLlib:2018}, offer support for training general purpose agents to a range of problem settings. However, these libraries are designed for optimizing fixed problem settings by varying agent parameters and do not offer support for variations in the problem or data source. %This is particularly important for language based solutions as the choice of encoding method used to transform language states is a crucial component that must be evaluated.

This requirement for the custom development of software for each RL application presents a challenge when posing new solutions. This also makes it more challenging to introduce new or variations of existing problem settings to evaluate the methodologies on. For example, it was observed that most new developments in Text Game settings were applied to either Jericho Suite of Games or TextWorld \citep{osborneSurveyTextGames2022}. Likewise, Gymnasium \citep{towers2024gymnasium} and Atari \citep{mnih2013PlayingAtariDeep} are commonly used as benchmarks to evaluate RL algorithms. 

% \begin{figure}[ht!]
%     \centering
%     \includegraphics[width=1\linewidth]{images/GUI/GUI_Example_Instruction.png}
%     \caption{GUI instruction input tab example.}
%     \label{fig:GUI}
% \end{figure}

These challenges are problematic for two key user groups: 1) local domain specialists that wish to apply new methodologies to their RL problem setting with variations in their data, environment model (including language specification) or language encoding method, and 2) domain stakeholders that wish to explore language-driven solutions, in particular based on large language models (LLMs), to reward-based environments without significant development work in RL.

Therefore, this work presents \textbf{elsciRL} as the first attempt at a general-purpose framework for the application of language solutions to a reward-based environment, including those originally defined without language. To our best knowledge, no library exists that formalizes an approach for introducing language to an RL problem setting~\citep{Osborne2024}. We demonstrate the reproducibility of our framework by applying an instruction following approach to a range of applications with minimal, problem specific requirements. All the code used in this work are published within the open-source Python library repository\footnote{\label{elsciRLfootnote}\url{https://github.com/pdfosborne/elsciRL}}. 

\begin{figure*}
    \centering 
    \includegraphics[width=1\linewidth]{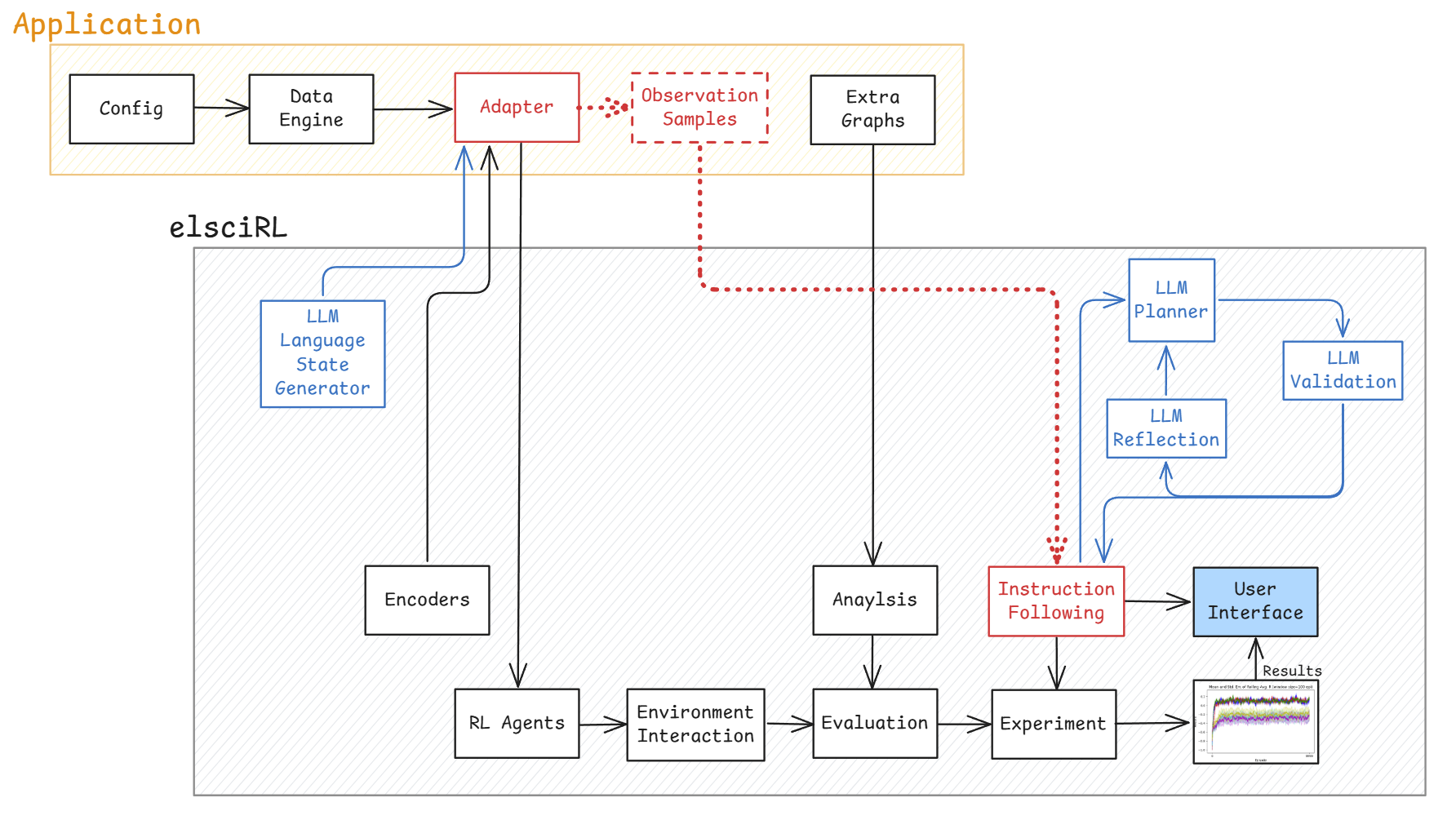}
    \caption{Overview of the \textbf{elsciRL} library, \textcolor{red}{red blocks} highlight the Language Adapter and Self-completing Instruction Following (LASIF) framework defined by \citep{Osborne2024}, \textcolor{blue}{blue blocks} highlight the methodological contributions of this work with LLMs.}
    \label{fig:elsciRL-overview}
\end{figure*}

To achieve this, we extend the \textit{HELIOS} framework~\citep{Osborne2024} that was used to apply the Language Adapter with Self-Completing Instruction Following (LASIF) methodology. Specifically, we recreate the complete software solution for the core library and add a Graphical User Interface (GUI) to enable user input and feedback. Furthermore, this work introduces a set of LLM based solutions into the LASIF framework to demonstrate the potential of the \textbf{elsciRL} software: 1) an LLM language adapter to generate text descriptions of numeric / symbolic states, 2) an LLM planner module to generate a set of instructions given a user input to complete the task, and 3) an LLM validation process that evaluates the unsupervised prediction of best match state for the completion of the current instruction.

A set of applications are used to evaluate each of the LLM methodologies highlighted in Figure \ref{fig:elsciRL-overview}. Specifically, two GridWorld based problems and a Maze problem defined by \citep{nlrl2024} where Table \ref{tab:applications} provides a summary of each and links to the application's repositories. For each application, we train and test a set of fixed RL agents and use \textit{elsciRL's} evaluation protocols to analyze the reward obtained after each episode which is defined as a complete interaction phase either to a terminal state (e.g. winning the game). 

We find that these LLM approaches can be used to improve the performance of a Q-learning and Deep-Q Network \citep{mnih2013PlayingAtariDeep} agent as shown in Table \ref{tab:results}. However, although we show that LLM adapters and self-completing instructions \textit{can} be used to improve results, further work is required to evaluate the use of these approaches and expand the number of application settings. It is for this reason that we are presenting this work as a software library so that obtaining more results can be achieved with minimal setup requirements.

\begin{table*}[h!t]
    \centering
    \resizebox{\textwidth}{!}{
    \begin{tabular}{p{0.11\linewidth}p{0.05\linewidth}p{0.625\linewidth}p{0.15\linewidth}}
        \hline
        \textbf{Application} &  & \textbf{Description} & \textbf{Link} \\
         \hline 
         {\small\textbf{Classroom}}  & {\small \citep{osborneApplyingReinforcementLearning2022}} & {\small A GridWorld task to recycle scrap paper with varying behavioured students, fixed starting and goal positions. If the paper is not recycled, a significant negative reward is given.} & {\small \href{https://github.com/pdfosborne/elsciRL-App-Classroom}{elsciRL-App-Classroom}} \\
         \hline 
         {\small\textbf{Gym FrozenLake}} & {\small \citep{towers2024gymnasium}} & {\small Help the Elf get the present without falling through ice, fixed starting and goal position. As defined by Gymnasium, no negative reward is given for falling into the ice.} & {\small \href{https://github.com/pdfosborne/elsciRL-App-GymFrozenLake}{elsciRL-App-GymFrozenLake}} \\ 
         \hline 
         % {\small\textbf{Sailing}} & {\small  \citep{Osborne2024}} & {\small Get the sailboat up the river against the wind without hitting the edges.} & {\tiny \url{https://github.com/pdfosborne/elsciRL-App-Sailing}} \\ 
         % \hline 
         % {\small\textbf{Chess}} & {\small \citep{Osborne2024}} & {\small A shortened game of Chess where the first capture wins.} & {\tiny \url{https://github.com/pdfosborne/elsciRL-App-Chess}} \\ 
         % \hline 
         {\small\textbf{Maze}} & {\small \citep{nlrl2024}} & {\small A set of mazes with language based positional descriptions, fixed starting and goal positions. All maze types (umaze, double-t, medium and large) are described with consistent language descriptions based on relative position to walls.} & {\small \href{https://github.com/pdfosborne/elsciRL-App-Maze}{elsciRL-App-Maze}}\\ 
         \hline
    \end{tabular}
    }
    \caption{Summarized description of the applications. \textit{Link} is the code repository identifier.}
    \label{tab:applications}
\end{table*}

% ----------------------
\newpage
\section{elsciRL}

The \textbf{elsciRL} library was developed following the results of \citep{Osborne2024} as a means to re-apply the LASIF framework to a range of applications. Furthermore, we intend that it can be used to improve the reproducibility and reduce the challenges of applying other language based solutions to RL environments. It achieves this by introducing a more complete structure to applying RL by: 1) generalizing the interaction process, 2) offering a set of evaluation protocols, 3) composing these into standardized experiments, and 4) enabling user input through a GUI. The software is summarized in Figure \ref{fig:elsciRL-overview} and examples of the GUI are provided in the images in Appendix \ref{sec:appendix-GUI-screenshots}.

%This is in contrast to prior libraries in RL such as StableBaselines3, etc (REF) that provide only the agent module. Typically, when evaluating new methodologies the rest of the application and hierarchy approaches are custom built making them hard to re-apply to new problems without refactoring. Now, with \textbf{elsciRL}, only the underlying data source, configuration and adapters that transform the observed state into the accepted form are required in the Application specification. This also means that when creating new solutions within the framework a complete set of applications are available with language specifications to be tested on.

%In order to use the LLM approaches you will need to have Ollama\footnote{\url{https://ollama.com/download}} installed with the model downloaded to be called. 

% ----------

\subsection{LLM Adapter}
\label{sec:adapter}

Adapters are defined in \citep{Osborne2024} as a means to transform observed states into a standardized form. They can also by used to add human supervision to the agent by giving context of the problem. Formally, they are applied using the function:

\begin{equation}
    f:s \in S \longmapsto o \in O
    \label{eqn:adapter}
\end{equation}

where $s$ is the environment's state and $o$ is the language observation. The difference between the size of the state and language space result in the partial observability of the transformation, i.e. \hl{}$|S| \geq |O|$). % ({\color{red} Couldn't a state be represented by more than one language observation?}).

%This \hl{could be a modification of the numeric representation to better align to other problems} ({\color{red} Needs to be more specific here}). 

%For example, card games are often represented independently so that, for example, Blackjack might be denoted with a list of just your hand and the dealer's visible card as an array $[card1, card2, dealer\_card]$. Likewise Poker could be your hand and the visible cards on the table as a \hl{a set of arrays} $[[card1, card2, ...], [table\_card1, table\_card2, ...]]$. This makes it impossible for the same agent to be trained on one Blackjack and applied to Poker and vice versa simply because the input dimensions are not consistent. Therefore, in this example an adapter could be used on each to transform to a consistent representation. The simplest being an array that represents the count of cards in the players hard, i.e. $\{Ace\_Hearts: 0, Two\_Hearts: 1, ...\} = [0,1, ...]$. This would therefore provide a consistent array with one dimension and 52 entries for all card problems but would only include the player's hard thereby losing information for some game types. This may be acceptable for Blackjack but would likely be detrimental to the performance of an agent on Poker given how impactful the cards on the table are to the outcome of the game. Significant consideration must be made to the impact of adding or removing context of the problem with these adapters.

For example, given the observation of a patient [gender=1, height=190cm, weight=70kg, build=‘slim’] it can first be mapped to [‘male’, ‘tall’, ‘normal weight’, ‘slim build’] and the concatenated output with connecting language terms to ‘A tall male of normal height and slim build’. The state space size will therefore be larger than the observation space size here because heights and weights are being grouped into collective terms thereby making it a partially observed view of the original state space.

Rule or logic based transformation approaches have been used in prior work as, for example, the Maze problem \citep{nlrl2024} has a state space defined numerically by the [$y$, $x$] positions. The authors use this to specify the language generation that transforms this information into a set of textual descriptions, such as \textit{`A wall is on your left, ...'} which can be provided to the LLM adapter alongside the numeric state representation.

To enhance the rule-based language generation, this work enables the use of an LLM to generate the language description by providing it the current state from the environment, previous actions and legal actions. A system level prompt is defined within the \textbf{elsciRL} library to specify the required format of the LLM's output. Furthermore, to save runtime, the LLM adapter will cache its generation so that it will use the first output generated for each state. The LLM adapter is called at the application level so the user can specify custom prompts to add further context to the LLM generation. The LLM adapter also includes usage of any available language transformers available within \textbf{elsciRL} without additional setup requirements.

%The LLM language generation is defined within the core \textbf{elsciRL} library, including the encoder selection, so that application specific adapters can simply call it when needed whilst still enabling adding context to the input.

% ----------
\subsection{LLM Instruction Following}
\label{sec:IF}

\subsubsection{Observation Samples}

In order to use an unsupervised instruction method we need to allow the approach to compare observed language descriptions with knowledge of where in the environment it was discovered. This can be completed during the training interaction of the agent alongside its exploration phase. 

However, to reduce the computation time during the agent's interaction we instead use a set of observed states that are obtained prior to the agent's training. This works well in applications where knowledge of the state space is expected but knowing which positions lead to good outcomes is not and ultimately what the agent must learn. For example, in a recommender system the set of all possible states may be defined based on the combination of items available. Obtaining these observed language states can be achieved by random exploration of the environment or predefined if the problem's state space can be specified. Both the underlying environment's state and the transformed language position as defined by Equation \ref{eqn:adapter} are stored.

\subsubsection{Self-Completing Instructions}
Given the set of observed states, instructions are self-completed using an unsupervised prediction as defined in \citep{Osborne2024}. For now, a cosine similarity measurement is used between the current instruction and the language produced by the adapter across all observed states to find a best match. To enable such an approach, the language from the instruction and adapter must be encoded using a consistent method. The choice of encoder can range from simplistic approaches such as Bag of Words to methods that incorporate pre-trained world knowledge such as text embeddings.

% ----------

Prior works of instructions or sub-goals required the completion to be defined by a human. For example, \citep{Tessler:2017} trained a hierarchy by defining long-term tasks as a connection of rooms, each with a specific task and a supervised completion state. Likewise, recent works of \citep{shu:interp-RL} and \citep{hu:nl-instr} required definition of the states that completed each instruction. Specifically, \citep{hu:nl-instr} labelled instruction completion based on having a human play the game. 

Given the already defined self-completing instruction following approach in \textbf{elsciRL}, we can simply enhance this process through the use of LLMs. Observed states are extracted based on the adapter used and therefore we can compare rule-based against LLM generated language. 

First, we input the human provided instruction to an LLM to break-down the objective into smaller steps. For each step, the unsupervised prediction method defined by \citep{Osborne2024} is used to find the most likely state matches. 

Prior to any human check we ask a separate LLM model to validate that the unsupervised prediction matches the expected instruction. The result should specify whether it believes the instruction is completed by the output best state match from instruction following search.

If not, a small negative reinforcement is provided to the matching algorithm that will adjust it's prediction as defined by \citep{Osborne2024}. Furthermore, we ask the LLM to \textit{`reflect'} on its instruction and to update this based on the language structure provided by the environment to improve its specification of the intended instruction (a.k.a \textit{`iterative refinement'} \citep{LLM_iter_refinement}). This process repeats until the LLM validation confirms a correct match or an arbitrary limit is reached. 

%REF https://arxiv.org/pdf/2303.17651

If confirmed as correct by user, the state (or states) which complete the instruction are used to specify a sub-goal within the environment and defined by the underlying numeric state representation. This means that instructions can be completed irrespective of any adapter used to train the agent and LLMs can be used within the instruction following without requiring an LLM agent. The agent is trained to complete instructions in the order they were given for a limited number of episodes. An additional reward for successfully reaching the predicted completion states is provided and is removed completely in testing.

The instruction results for this work are summarised in Table \ref{tab:results-instruction} as well as being published within each application repository (see Table \ref{tab:applications}). This includes the original user input, the LLM's generated sub-instructions and then the resultant best match state from the environment for each. These instructions are then used to train and test a set of RL agents to produce the results in Table \ref{tab:results} and Appendix \ref{sec:appendix-graphical-results}.

\begin{table*}[h!]
    \hspace{-.6cm}
    \resizebox{1.05\textwidth}{!}{
    \begin{tabular}{c|p{0.34\textwidth}|p{0.66\textwidth}|c}
    {\small\textbf{Application}}  & {\small\textbf{User Input}} & {\small\textbf{LLM Instructions}} & {\small\textbf{State}} \\
    \hline
      \multirow{4}{*}{{\small Classroom}}  & {\small Pass the paper to the teacher without it going to the punk student, you cannot} & {\small 1. Locate and move away from the punk student until you are no longer in their vicinity} & {\small[1,3]}\\
      & {\small move students so must avoid him by going the long way round the classroom} & {\small 2. Hand over the paper to the teacher while avoiding contact with the punk student} & {\small[3,3]}\\
      \hline
      
      \multirow{2}{*}{{\small FrozenLake}}  & {\small Help the elf reach the present in the} & {\small 1. Move north from the starting position until reaching the present} & {\small 3}\\
      & {\small Gymnasium Frozen Lake environment} & {\small 2. Verify the present has been successfully accessed by stepping on it} & {\small 15}\\
      \hline
      
      \multirow{4}{*}{{\small UMaze}} & {\small Help the agent complete the umaze maze problem by providing guidance} & {\small 1. Move towards the nearest wall and take one step along and take one step along it...} & {\small[2,1]}\\
      & {\small in the form of two exact [y,x] sub-goals positions it should reach} & {\small 2. If the agent reaches a dead end move back to the previous position and try again...} & {\small[3,3]}\\
      \hline

      \multirow{3}{*}{{\small Double-T}} & {\small Help the agent complete the double-t-maze maze problem by providing...} & {\small 1. Choose the correct direction to take on the outer track T to avoid obstacles...} & {\small[7,6]}\\
      & & {\small 2. Move to the end of one of the tracks} & {\small[7,6]}\\

    \end{tabular}
    }
    \caption{Instruction results with best match environment state from the unsupervised prediction.}
    \label{tab:results-instruction}
\end{table*}

% NOT SURE ABOUT SELF EVALUATION WITH LLM AT THIS STAGE. THE VALIDATION IS NOT CURRENTLY EFFECTING THE UNSUPERVISED PREDICTION AT THIS STAGE AND THE LLM CANNOT PROVIDE CONFIDENCE FOR ITS RESULT IN A CONSISTENT METRIC. IF IT ISN'T CONFIDENT IT MATCHES THEN NOTHING HAPPENS ANYWAY...

% ADD ANALYSIS ON SIM SCORE FOR MATCHING TO EVALUATION AND CORRECT VS INCORRECT VALIDATED NUMBERS

% NEED TO SPECIFY A STANDARD SET OF INPUT INSTRUCTIONS FOR EVAL

\subsection{Installation, Setup and Basic Usage}

The \textbf{elsciRL} library can be installed by following the instructions on the GitHub repository or with:

\begin{verbatim}
    pip install elsciRL
\end{verbatim}

Once installed, run the Python script:

\begin{verbatim}
    from elsciRL import App
    App.run()
\end{verbatim}

This will run the GUI which is a Flask app that can be viewed on a web browser with the local-host address: \textit{http://127.0.0.1:5000}. The app allows you to select from the application registry to apply with the instruction following approach and then run the RL agents.

Previews of the GUI are shown in Appendix \ref{sec:appendix-GUI-screenshots} Figures \ref{fig:GUI-Example-Application} - \ref{fig:GUI-Example-Train}. To run an experiment, the user needs to complete the following steps:

\begin{enumerate}
    \item Select an application, this will load a preview of the problem and allow you to select a sub-problem configuration (e.g. different maze types). The user will also be able to select the observed states data used for the unsupervised instruction following step.
    \item Select the training and testing parameters, agent and adapter combinations and agent parameters. Published configurations can be imported or custom options can be exported/imported as needed.
    \item (Optional) Provide instruction inputs, by default this will directly match the user's input against the environment's observed states. The user will be presented with the predicted best match to confirm. They can provide multiple instructions per input if separated by a new line and after confirming can use the `New Instruction' button to add more inputs. If LLM planner is enabled then an LLM will attempt to improve your input based on the configuration options selected. Published instructions can also be imported including those given in Table \ref{tab:results-instruction}.
    \item Once all configuration options and instructions are finalized, the user can run the experiment. This will complete the train/testing phase for all selected agents and adapter combinations, with and without the input instructions. When the experiment is complete all figures are displayed on the results tab and saved in a local directory.
\end{enumerate}
    
% ----------------------

% ----------

\begin{table*}[ht!]
    \centering
    \resizebox{0.96\textwidth}{!}{
    \begin{tabular}{|cc||c|ccc||c|ccc|}
        \hline
          & & \multicolumn{4}{c||}{\textbf{Q-Learning Agent}} & \multicolumn{4}{|c|}{\textbf{DQN Agent}} \\
          {\small\textbf{Application}} & {\small\textit{Base Adpt.}} &  & {\small LLM Adpt}. & {\small LLM IF} & {Comb.} &  & {\small LLM Adpt.} & {\small LLM IF} & {\small Comb.} \\
         \hline
         \multirow{2}{*}{{\small\textbf{Classroom}}} & {\small \textit{Numeric}} & \textit{-0.48} & \multirow{2}{*}{-0.78} & \textbf{-0.33} & \multirow{2}{*}{-0.77} & \textit{-0.49} & \multirow{2}{*}{-0.69} & \textbf{-0.28}  & \multirow{2}{*}{-0.64} \\
         & {\small\textit{Language}} & \textit{0.12}  &  & \textbf{0.13} &  & \textit{0.05} &  & \textbf{0.16} &  \\
        \hline
        \multirow{2}{*}{{\small\textbf{FrozenLake}}} & {\small\textit{Numeric}} &  \textbf{\textit{0.13}}& \multirow{2}{*}{0.01} & 0.00 & \multirow{2}{*}{0.02} & \textit{0.00} & \multirow{2}{*}{0.00} & 0.00 & \multirow{2}{*}{\textbf{0.10}} \\
         & {\small\textit{Language}} & \textit{0.00}  &  & 0.00 &  & \textit{0.00} &  & 0.00 & \\
        \hline
        \multirow{2}{*}{{\small\textbf{Maze}}} & {\small\textit{Umaze}} & \textbf{\textit{1.00}} & 0.96 & -0.10 & 0.96 & \textit{-0.10} & -0.10 & -0.10 & \textbf{0.12} \\
         & {\small\textit{Double-t maze}} & \textbf{\textit{0.99}} & 0.22 & -0.10 & 0.23 & \textbf{\textit{0.99}} & -0.10 & \textbf{0.99} & 0.99 \\
        % & {\small\textit{Medium maze}} & - & - & - & - & - & - & - & - \\
        % & {\small\textit{Large maze}} & - & - & - & - & - & - & - & - \\
        % \hline
        % \multirow{ 2}{*}{{\small\textbf{Sailing}}} & {\small\textit{Numeric}} & - & - & - & - & - & - & - & - \\
        %  & {\small\textit{Language}} & - & - & - & - & - & - & - & - \\
        % \hline
        % \multirow{ 3}{*}{{\small\textbf{Chess}}} & {\small\textit{Numeric Board}} & - & - & - & - & - & - & - & - \\
        % & {\small\textit{Piece Counter}} & - & - & - & - & - & - & - & - \\
        %  & {\small\textit{Active Pieces Language}} & - & - & - & - & - & - & - & - \\
         \hline
    \end{tabular}
    }
    \caption{\textbf{Testing Results:} The best agent from 10 training repeats and 10,000 episodes is selected and applied with fixed policy for 1,000 episodes and repeats another 10 times. \textit{Baseline} are non-LLM adapters without any instructions used with scores given in the first column for each agent. For each, two instructions are produced by Llama 3.2 for the instruction following (IF) method.}
    \label{tab:results}
\end{table*}

\section{Evaluation}

Typically, RL agents are evaluated on a static problem setting against a set of baseline comparisons. In our work, we flip this setup such that the agents are fixed and what we are instead evaluating is a change to the specification of the problem setting. First, adapters transform the the state space and then the instruction following approach provide the agents with additional reward in training for reaching the desired outcome. 

The \textbf{elsciRL} library provides a set of evaluation protocols that analyse the performance of training agents by the obtained reward and are used to produce the tabular and graphical results (see Appendix \ref{sec:appendix-graphical-results} for examples). Furthermore, all agent parameters used in this work are fixed and are available as part of the \textbf{elsciRL} library within the \textit{`Published Experiments'}\footnote{\href{https://github.com/pdfosborne/elsciRL/blob/main/elsciRL/published\_experiments/osborne\_2025.py}{github.com/pdfosborne/elsciRL/blob/main/elsciRL/\\published\_experiments/osborne\_2025.py}} directory so that they can be used in future work as a baseline.

%An example is show in figure \ref{fig:variance-training-example} where three graphs represent the reward obtained and the fourth shows the time taken per episode.

Testing performance is measured by applying the trained agent with a fixed policy. In this work we use a Single Setting evaluation so that the trained agent is applied to the same environment without additional rewards (i.e. without instruction following reward guidance). We use two agents, a Q-Learning tabular and a Deep-Q Network agent as defined by \citep{mnih2013PlayingAtariDeep}. Both agents are trained for 10,000 episodes with 10 independent repeats, the best agent from training is selected and then testing for 1,000 episodes with another 10 repeats.

A tabular agent is useful for evaluating the impact of partial observability introduced by each adapter because it simply treats acts on each unique state. Therefore, when multiple states are transformed to the same language description then the agent will make the same action across all state. This will be the same challenge for all agents except the Q-Learning agent has no mechanism to transfer its decision making across similar states making it a suitable control agent to only evaluate the impact of the adapter's partial observability. 

Alternatively, to evaluate the impact of language as a mechanism to transfer knowledge between states we use a Deep-Q Network agent. The Deep-Q Network allows for the features within each state to be used as inputs to a neural network thereby allowing similar states to adopt a similar policy.

We evaluate the performance change of training and testing the agents with the LLM adapter and instruction following approaches defined previously in Sections \ref{sec:adapter} and \ref{sec:IF} respectively. In all LLM approaches of this paper we use Llama3.2 \citep{llama3herdmodels}, \citep{llama32} and encode language using the \textit{MiniLMv6} model via the SentenceTransformers library \citep{wang:MiniLM}. We obtain results for the set of applications given in Table \ref{tab:applications}. To complete the instruction following method, observed states data is obtained using the LLM adapter with Llama3.2 for each problem setting.

% \subsection{Applications}

% A set of applications have been updated with our LLM adapters and, as they are designed within the \textbf{elsciRL} library, can be run individually to reproduce all our results. Each application can be defined independently with varying authors provided they are in required form and then imported into an experiment. Currently, this is achieved by publishing each application as a separate public GitHub repository. The applications used in this work are provided in Table \ref{tab:applications}.

% The applications must include: 1) the data sourced defined as a reward based environment and, 2) adapters to specify the state form. Additionally, problem specific figures can be included and sampled observed data is collected in advance for the unsupervised instruction following method.

% ----------------------
\section{Results and Discussion}

Table \ref{tab:results-instruction} show the LLM's instruction plan given the user input and the resultant best match state from the unsupervised prediction. Due to the simplicity of the environments we can confirm that these are valid and the predicted match states achieve the intended outcomes. Specifically,

\noindent
\textbf{Classroom:} [4,1] $\rightarrow$ [1,3] $\rightarrow$ [3,3] would avoid the punk student (see Figure \ref{fig:app-classroom}) and go around the room to reach the teacher at [3,3].

\noindent
\textbf{FrozenLake:} going \textit{`North'} from the starting position isn't directly possible (see Figure \ref{fig:app-frozenlake}) but as \textit{`the lake is slippery so the player may move perpendicular to the intended direction'} \citep{towers2024gymnasium} then this results in a movement left or right, therefore 0 $\rightarrow$ 3 $\rightarrow$ 15 would be expected from these instructions to get the present at state 15.

\noindent
\textbf{Maze:} instructions in both cases are valid but do not provide specific state positions, however, due to the simplicity of the maze these positions are on-route to the goal states.\\[-6pt]

Table \ref{tab:results} show the testing results when applying a Q-Learning tabular agent and a Deep-Q Network neural agent to each application. A complete set of graphical results are provided in Appendix \ref{sec:appendix-graphical-results} Figures \ref{fig:results-classroom-train-instr} - \ref{fig:results-double-t-maze-test-noinstr}. In summary, we find that the graphical results show that the instruction following approach impacts the reward obtained in early episodes during training. Likewise, testing results in Table \ref{tab:results} indicate the instruction following approach can improve of the agent's performance. Furthermore, the Deep-Q network agent performs worse than the tabular Q-learning agent due to the simplicity of these problem settings. However, the instructions can cause the agent to get stuck at the short-term objective instead of helping to achieve a long-term goal as shown for the Maze's Q-Learning results. 

We also find that the LLM adapter's language does not support improvements in most cases. However, as the testing process selects the best agent from training, it is possible that in some cases none of the training agents overcame poor exploration due to randomness in the environment. Although 5-10 repeated training samples is standard in RL \citep{Agarwal:Evaluate-RL}, this may still be insufficient for proving definitive results. It may also be worthwhile using all training agents in testing, as this would greatly increase the variation of the testing results.

\section{Conclusion}

This work introduces \textbf{elsciRL}, an open-source Python library to facilitate the application of language solutions on reinforcement learning problems. We highlight the impact of this work as currently no framework exists for the development of language solutions in reinforcement learning. We intend this work to be used to accelerate the research of two key users groups: 1) domain specialists that wish to apply language based solutions to their problem settings, and 2) researchers that wish to evaluate language solutions on a range of reinforcement learning problems.

We outline how new applications can be designed using minimal, problem specific setup. In addition, we demonstrate how an instruction following approach can be applied to a small set of applications. We find that although results show promise, further evaluation must be conducted including more instructions and applications.

Opportunities for future work including evaluation of alternatives for the: 1) agent types (e.g. PPO or LLM agents), 2) language transformers, 3) LLMs models, and 4) unsupervised instruction completion methods. Lastly, generalisability could be evaluated within and across problems.

% \subsection{Future Work}

% - Policy gradient agents
% - LLM Agents
% - Language Encoders beyond MiniLMv6
% - Applications
% - Generalisability with evaluation
% - Use of visual LLMs 

% \section*{Limitations}
% EMNLP 2023 requires all submissions to have a section titled ``Limitations'', for discussing the limitations of the paper as a complement to the discussion of strengths in the main text. This section should occur after the conclusion, but before the references. It will not count towards the page limit.  

% The discussion of limitations is mandatory. Papers without a limitation section will be desk-rejected without review.
% ARR-reviewed papers that did not include ``Limitations'' section in their prior submission, should submit a PDF with such a section together with their EMNLP 2023 submission.

% ----
% \bibliographystyle{acl_natbib}
\bibliography{custom}

\begin{thebibliography}{18}
\expandafter\ifx\csname natexlab\endcsname\relax\def\natexlab#1{#1}\fi

\bibitem[{Agarwal et~al.(2021)Agarwal, Schwarzer, Castro, Courville, and Bellemare}]{Agarwal:Evaluate-RL}
Rishabh Agarwal, Max Schwarzer, Pablo~Samuel Castro, Aaron~C. Courville, and Marc~G. Bellemare. 2021.
\newblock Deep reinforcement learning at the edge of the statistical precipice.
\newblock \emph{Advances in Neural Information Processing Systems (NeuIPS)}.

\bibitem[{D'Eramo et~al.(2021)D'Eramo, Tateo, Bonarini, Restelli, and Peters}]{MushroomRL:2021}
Carlo D'Eramo, Davide Tateo, Andrea Bonarini, Marcello Restelli, and Jan Peters. 2021.
\newblock \href {http://jmlr.org/papers/v22/18-056.html} {Mushroomrl: Simplifying reinforcement learning research}.
\newblock \emph{Journal of Machine Learning Research}, 22(131):1--5.

\bibitem[{Feng et~al.(2024)Feng, Wan, Fu, Liu, Yang, Koushik, Hu, Wen, and Wang}]{nlrl2024}
Xidong Feng, Ziyu Wan, Haotian Fu, Bo~Liu, Mengyue Yang, Girish~A. Koushik, Zhiyuan Hu, Ying Wen, and Jun Wang. 2024.
\newblock \href {http://arxiv.org/abs/2411.14251} {Natural language reinforcement learning}.

\bibitem[{Grattafiori et~al.(2024)Grattafiori, Dubey, Jauhri, and et~al.}]{llama3herdmodels}
Aaron Grattafiori, Abhimanyu Dubey, Abhinav Jauhri, and et~al. 2024.
\newblock \href {http://arxiv.org/abs/2407.21783} {The llama 3 herd of models}.

\bibitem[{Hu et~al.(2019)Hu, Yarats, Gong, Tian, and Lewis}]{hu:nl-instr}
Hengyuan Hu, Denis Yarats, Qucheng Gong, Yuandong Tian, and Mike Lewis. 2019.
\newblock Hierarchical decision making by generating and following natural language instructions.
\newblock \emph{33rd Conference on Neural Information Processing Systems (NeurIPS 2019)}, pages 10025--10034.

\bibitem[{Huang et~al.(2022)Huang, Dossa, Ye, Braga, Chakraborty, Mehta, and Araújo}]{huang2022cleanrl}
Shengyi Huang, Rousslan Fernand~Julien Dossa, Chang Ye, Jeff Braga, Dipam Chakraborty, Kinal Mehta, and João~G.M. Araújo. 2022.
\newblock \href {http://jmlr.org/papers/v23/21-1342.html} {Cleanrl: High-quality single-file implementations of deep reinforcement learning algorithms}.
\newblock \emph{Journal of Machine Learning Research}, 23(274):1--18.

\bibitem[{Liang et~al.(2018)Liang, Liaw, Nishihara, Moritz, Fox, Goldberg, Gonzalez, Jordan, and Stoica}]{RLlib:2018}
Eric Liang, Richard Liaw, Robert Nishihara, Philipp Moritz, Roy Fox, Ken Goldberg, Joseph~E. Gonzalez, Michael~I. Jordan, and Ion Stoica. 2018.
\newblock {RLlib}: Abstractions for distributed reinforcement learning.
\newblock In \emph{International Conference on Machine Learning ({ICML})}.

\bibitem[{Madaan et~al.(2023)Madaan, Tandon, Gupta, Hallinan, Gao, Wiegreffe, Alon, Dziri, Prabhumoye, Yang, Gupta, Majumder, Hermann, Welleck, Yazdanbakhsh, and Clark}]{LLM_iter_refinement}
Aman Madaan, Niket Tandon, Prakhar Gupta, Skyler Hallinan, Luyu Gao, Sarah Wiegreffe, Uri Alon, Nouha Dziri, Shrimai Prabhumoye, Yiming Yang, Shashank Gupta, Bodhisattwa~Prasad Majumder, Katherine Hermann, Sean Welleck, Amir Yazdanbakhsh, and Peter Clark. 2023.
\newblock Self-refine: iterative refinement with self-feedback.
\newblock In \emph{Proceedings of the 37th International Conference on Neural Information Processing Systems}, NIPS '23, Red Hook, NY, USA. Curran Associates Inc.

\bibitem[{Meta(2024)}]{llama32}
Meta. 2024.
\newblock \href {https://ollama.com/library/llama3.2} {Ollama/llama3.2}.
\newblock Web page.

\bibitem[{Mnih et~al.(2013)Mnih, Kavukcuoglu, Silver, Graves, Antonoglou, Wierstra, and Riedmiller}]{mnih2013PlayingAtariDeep}
Volodymyr Mnih, Koray Kavukcuoglu, David Silver, Alex Graves, Ioannis Antonoglou, Daan Wierstra, and Martin Riedmiller. 2013.
\newblock \href {https://doi.org/10.48550/arXiv.1312.5602} {Playing {{Atari}} with {{Deep Reinforcement Learning}}}.

\bibitem[{Osborne(2024)}]{Osborne2024}
Philip Osborne. 2024.
\newblock \emph{Improving Real-World Reinforcement Learning by Self Completing Human Instructions on Rule Defined Language}.
\newblock Phd thesis, The University of Manchester, Manchester, UK.

\bibitem[{Osborne et~al.(2022{\natexlab{a}})Osborne, N{\~o}mm, and Freitas}]{osborneSurveyTextGames2022}
Philip Osborne, Heido N{\~o}mm, and Andr{\'e} Freitas. 2022{\natexlab{a}}.
\newblock \href {https://doi.org/10.1162/tacl_a_00495} {A {{Survey}} of {{Text Games}} for {{Reinforcement Learning Informed}} by {{Natural Language}}}.
\newblock \emph{Transactions of the Association for Computational Linguistics}, 10:873--887.

\bibitem[{Osborne et~al.(2022{\natexlab{b}})Osborne, Singh, and Taylor}]{osborneApplyingReinforcementLearning2022}
Philip Osborne, Kajal Singh, and Matthew~E. Taylor. 2022{\natexlab{b}}.
\newblock \emph{Applying {{Reinforcement Learning}} on {{Real-World Data}} with {{Practical Examples}} in {{Python}}}.
\newblock Synthesis {{Lectures}} on {{Artificial Intelligence}} and {{Machine Learning}}. Springer International Publishing, Cham.

\bibitem[{Raffin et~al.(2021)Raffin, Hill, Gleave, Kanervisto, Ernestus, and Dormann}]{stable-baselines3}
Antonin Raffin, Ashley Hill, Adam Gleave, Anssi Kanervisto, Maximilian Ernestus, and Noah Dormann. 2021.
\newblock \href {http://jmlr.org/papers/v22/20-1364.html} {Stable-baselines3: Reliable reinforcement learning implementations}.
\newblock \emph{Journal of Machine Learning Research}, 22(268):1--8.

\bibitem[{Shu et~al.(2018)Shu, Xiong, and Socher}]{shu:interp-RL}
Tianmin Shu, Caiming Xiong, and Richard Socher. 2018.
\newblock Hierarchical and interpretable skill acquisition in multi-task reinforcement learning.
\newblock \emph{International Conference on Learning Representations (ICLR)}, 6.

\bibitem[{Tessler et~al.(2017)Tessler, Givony, Zahavy, Mankowitz, and Mannor}]{Tessler:2017}
Chen Tessler, Shahar Givony, Tom Zahavy, Daniel~J. Mankowitz, and Shie Mannor. 2017.
\newblock A deep hierarchical approach to lifelong learning in minecraft.
\newblock In \emph{Proceedings of the Thirty-First AAAI Conference on Artificial Intelligence}, AAAI'17, page 1553–1561. AAAI Press.

\bibitem[{Towers et~al.(2024)Towers, Kwiatkowski, Terry, Balis, De~Cola, Deleu, Goul{\~a}o, Kallinteris, Krimmel, KG et~al.}]{towers2024gymnasium}
Mark Towers, Ariel Kwiatkowski, Jordan Terry, John~U Balis, Gianluca De~Cola, Tristan Deleu, Manuel Goul{\~a}o, Andreas Kallinteris, Markus Krimmel, Arjun KG, et~al. 2024.
\newblock Gymnasium: A standard interface for reinforcement learning environments.
\newblock \emph{arXiv preprint arXiv:2407.17032}.

\bibitem[{Wang et~al.(2020)Wang, Wei, Dong, Bao, Yang, and Zhou}]{wang:MiniLM}
Wenhui Wang, Furu Wei, Li~Dong, Hangbo Bao, Nan Yang, and Ming Zhou. 2020.
\newblock \href {http://arxiv.org/abs/2002.10957} {Minilm: Deep self-attention distillation for task-agnostic compression of pre-trained transformers}.

\end{thebibliography}

\newpage
\appendix
\section{GUI App Screenshots}
\label{sec:appendix-GUI-screenshots}

% -------------------------------------------
\noindent%
\begin{minipage}{\textwidth}% to keep image and caption on one page
\makebox[\linewidth]{%        to center the image
  \includegraphics[width=1\textwidth]{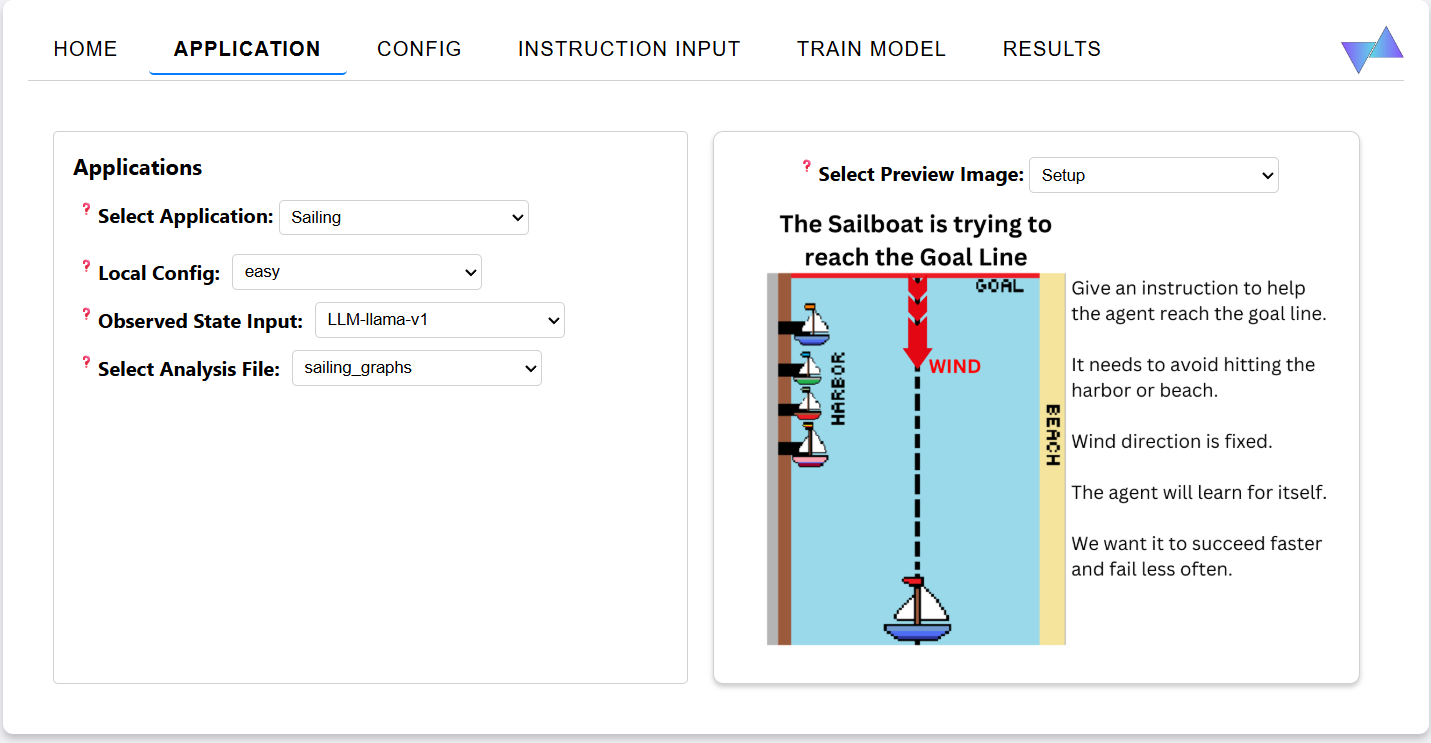}}
\captionof{figure}{GUI Application Selection Tab.}\label{fig:GUI-Example-Application}%      only if needed  
\end{minipage}

~\\

% \begin{figure*}[ht!]
%     \includegraphics[width=1\linewidth]{images/GUI/GUI_Example_Application.png}
%     \caption{GUI Application Selection Tab.}
%     \label{fig:GUI-Example-Application}
% \end{figure*}

\noindent%
\begin{minipage}{\textwidth}% to keep image and caption on one page
\makebox[\linewidth]{%        to center the image
  \includegraphics[width=1\textwidth]{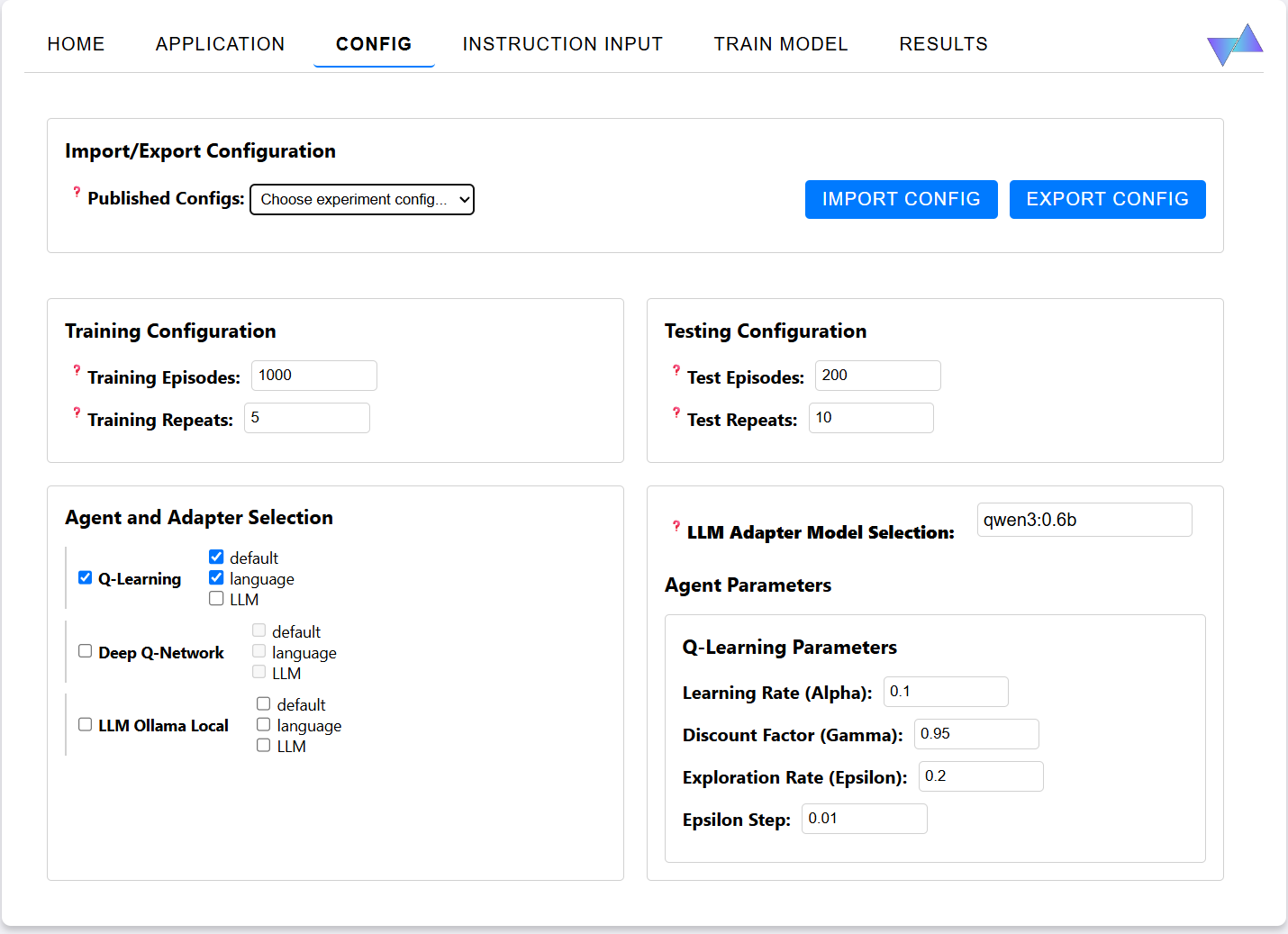}}
\captionof{figure}{GUI Agent Configuration Selection Tab.}\label{fig:GUI-Example-Config}%      only if needed  
\end{minipage}

% \begin{figure*}
%     \includegraphics[width=1\linewidth]{images/GUI/GUI_Example_Config.png}
%     \caption{GUI Agent Configuration Selection Tab.}
%     \label{fig:GUI-Example-Config}
% \end{figure*}

\begin{figure*}
    \includegraphics[width=1\linewidth]{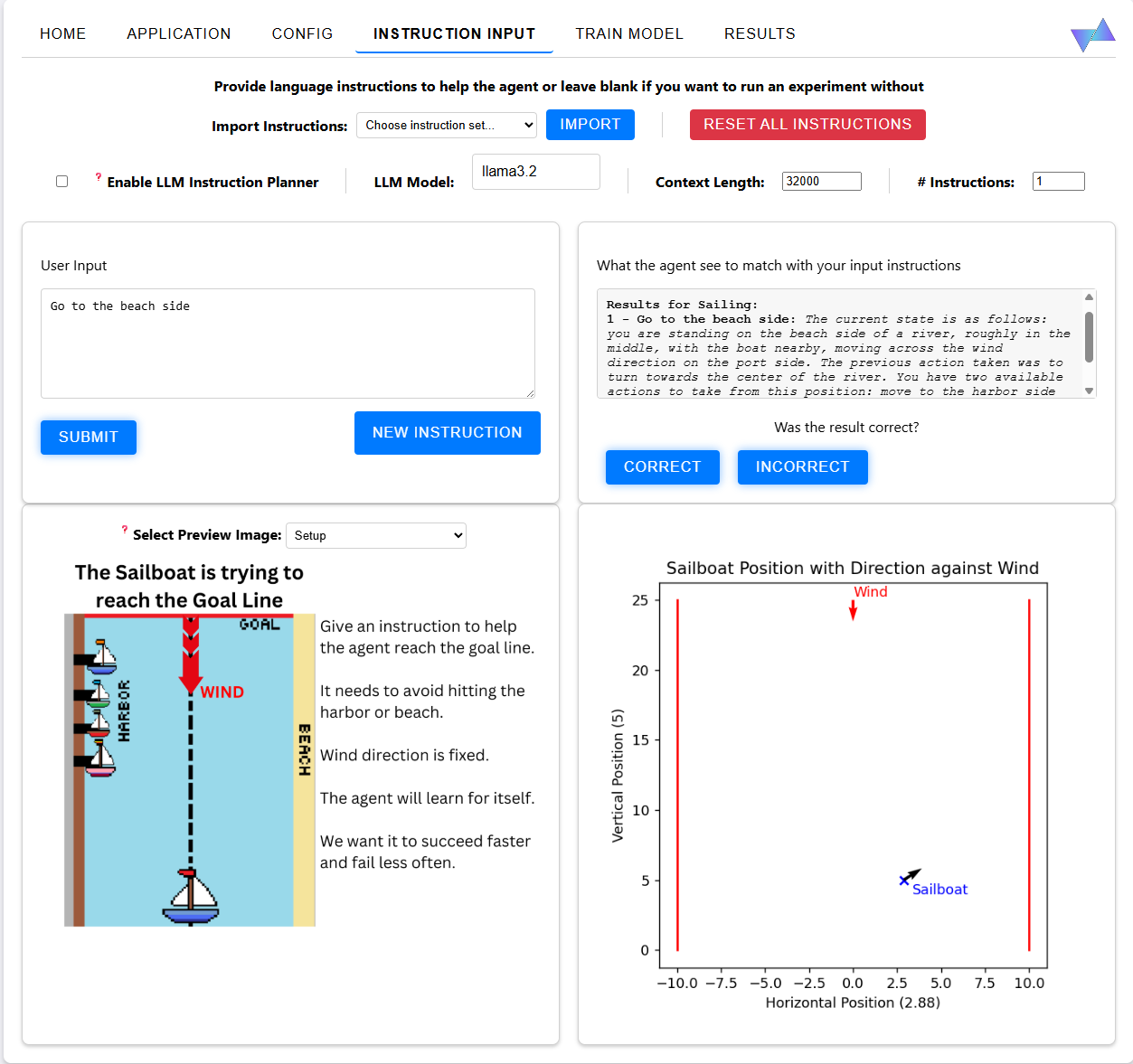}
    \caption{GUI Instruction Following Input Tab.}
    \label{fig:GUI-Example-Instruction}
\end{figure*}

\begin{figure*}
    \includegraphics[width=1\linewidth]{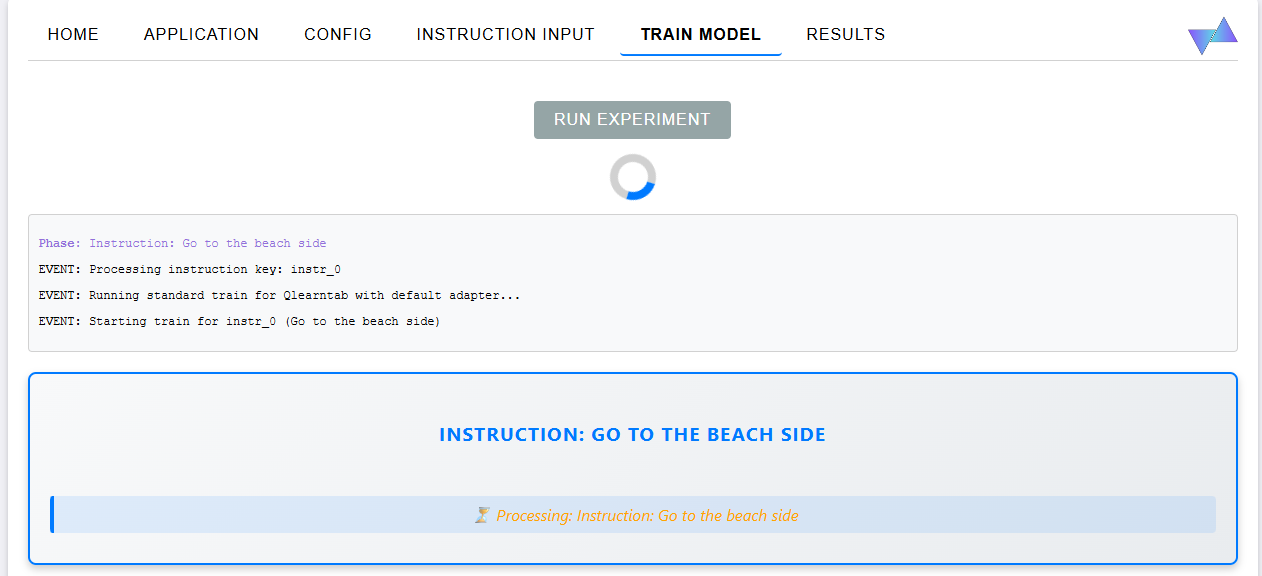}
    \caption{GUI Run Experiment Tab.}
    \label{fig:GUI-Example-Train}
\end{figure*}

% -------------------------------------------
\newpage\phantom{blankspace}
\newpage\phantom{blankspace}
\section{Application Environments}
\label{sec:appendix-application-screenshots}

\begin{figure}[h!]
    \centering
    \includegraphics[width=0.9\linewidth]{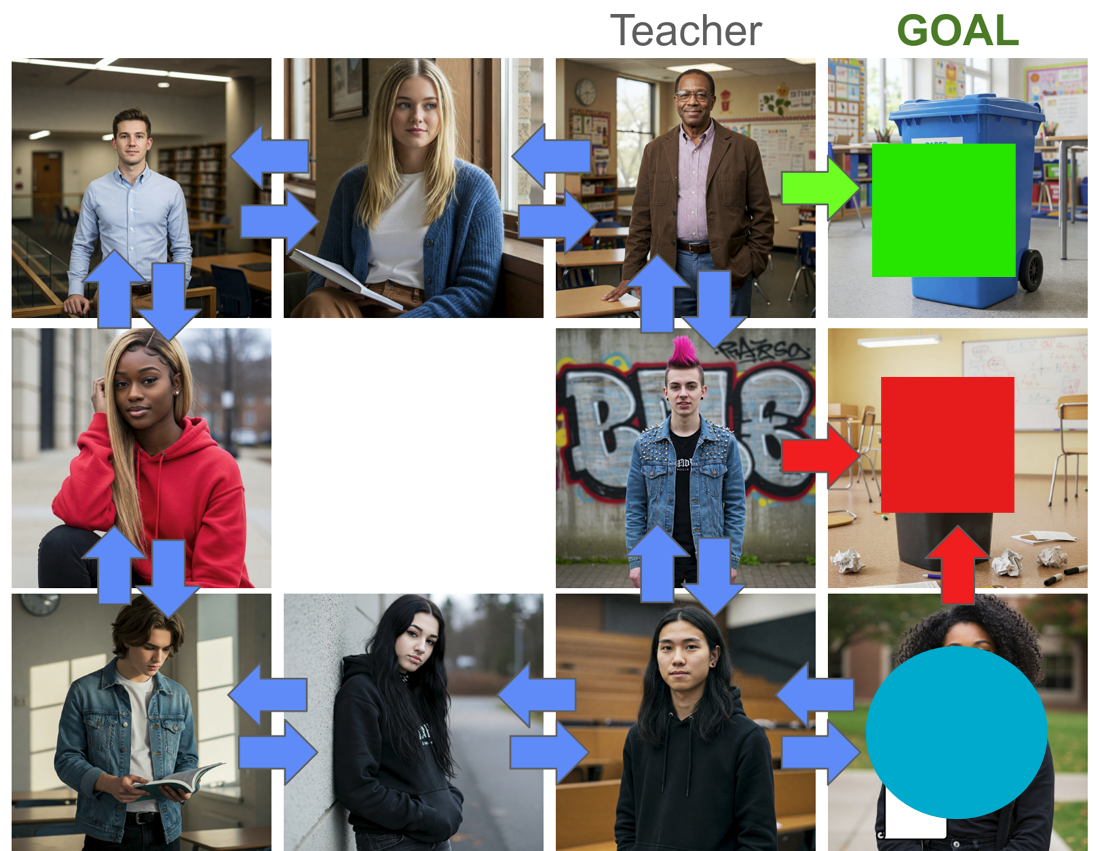}
    \caption{Classroom environment, start position fixed at [4,1], goal at [4,3] and negative outcome at [4,2].}
    \label{fig:app-classroom}
\end{figure}

\begin{figure}[h!]
    \centering
    \includegraphics[width=0.75\linewidth]{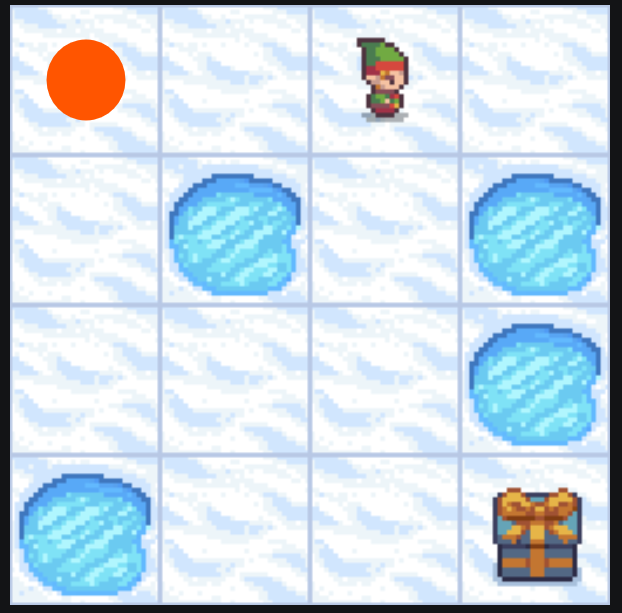}
    \caption{Gym-FrozenLake environment, start position fixed at 0, goal at 15.}
    \label{fig:app-frozenlake}
\end{figure}

\begin{figure}[h!]
    \centering
    \includegraphics[width=0.75\linewidth]{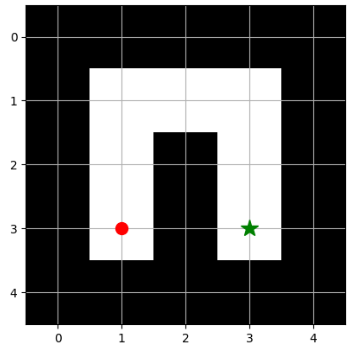}
    \caption{Umaze environment, start position fixed at [3,1], goal at [3,3], note the environment uses [y,x] coordinate geometry.}
    \label{fig:app-umaze_1}
\end{figure}

\begin{figure}[h!]
    \centering
    \includegraphics[width=0.75\linewidth]{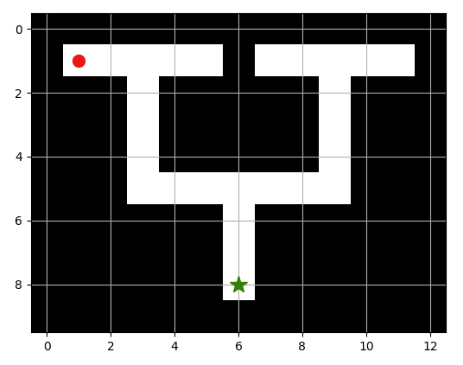}
    \caption{Double-T maze environment, start position fixed at [1,1], goal at [8,6], note the environment uses [y,x] coordinate geometry.}
    \label{fig:app-umaze_2}
\end{figure}

% -------------------------------------------

\newpage
\section{Graphical Results}
\label{sec:appendix-graphical-results}

The following provide an index lookup for each figure.

\begin{figure*}[h!]
    \centering
    \includegraphics[width=1\linewidth]{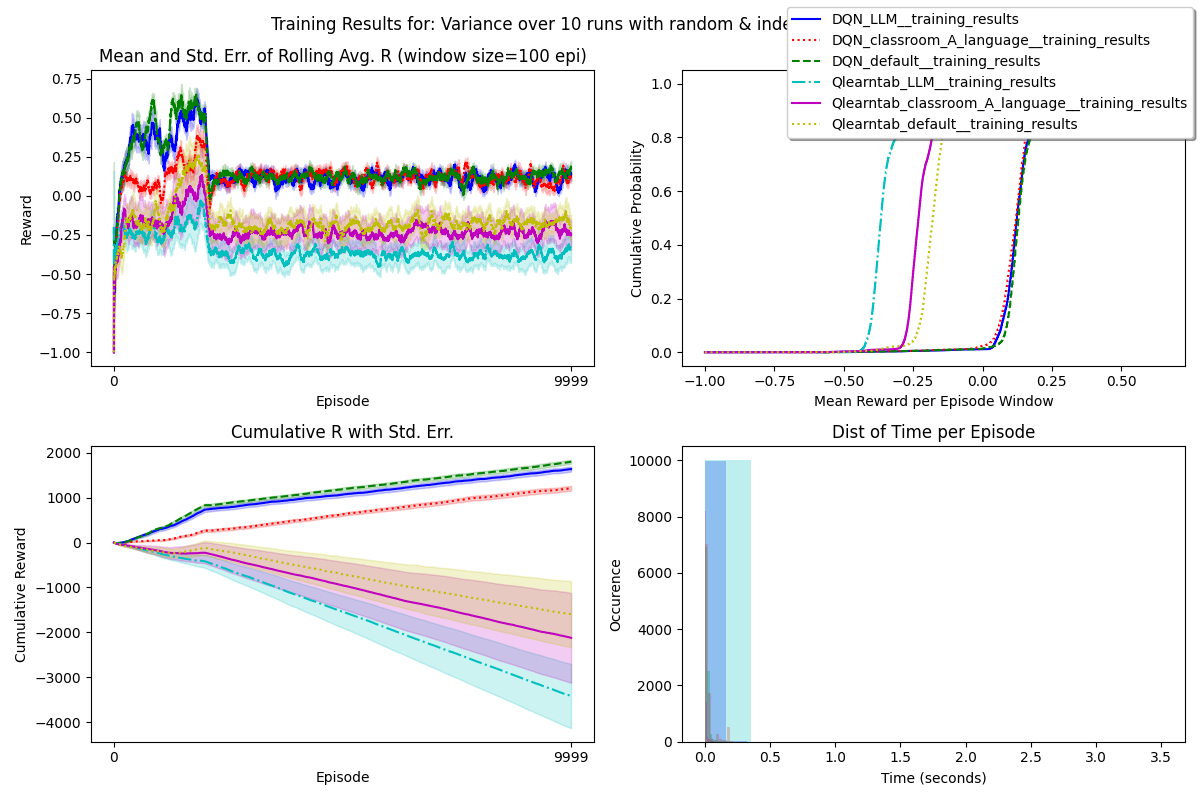}
    \caption{Classroom environment Instruction training results.}
    \label{fig:results-classroom-train-instr}
\end{figure*}

\begin{figure*}[h!]
    \centering
    \includegraphics[width=1\linewidth]{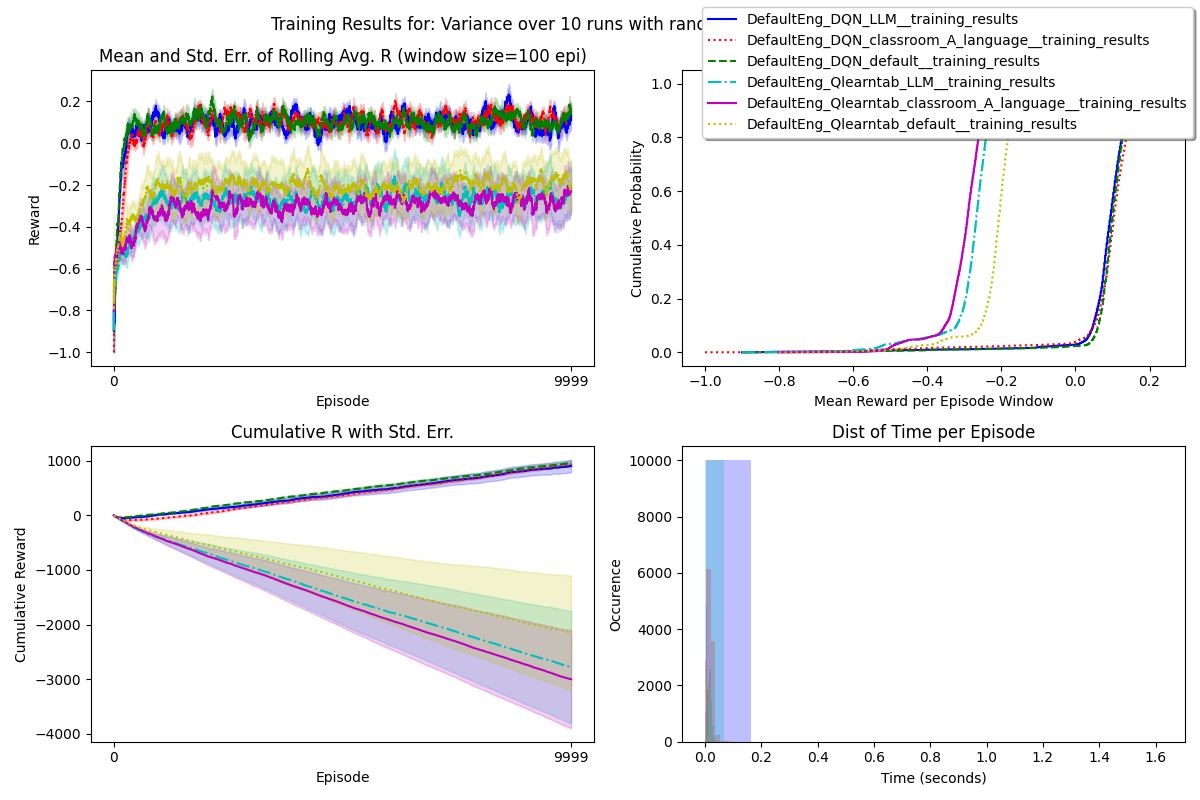}
    \caption{Classroom environment NO instruction training results.}
    \label{fig:results-classroom-train-noinstr}
\end{figure*}

\begin{figure*}[h!]
    \centering
    \includegraphics[width=1\linewidth]{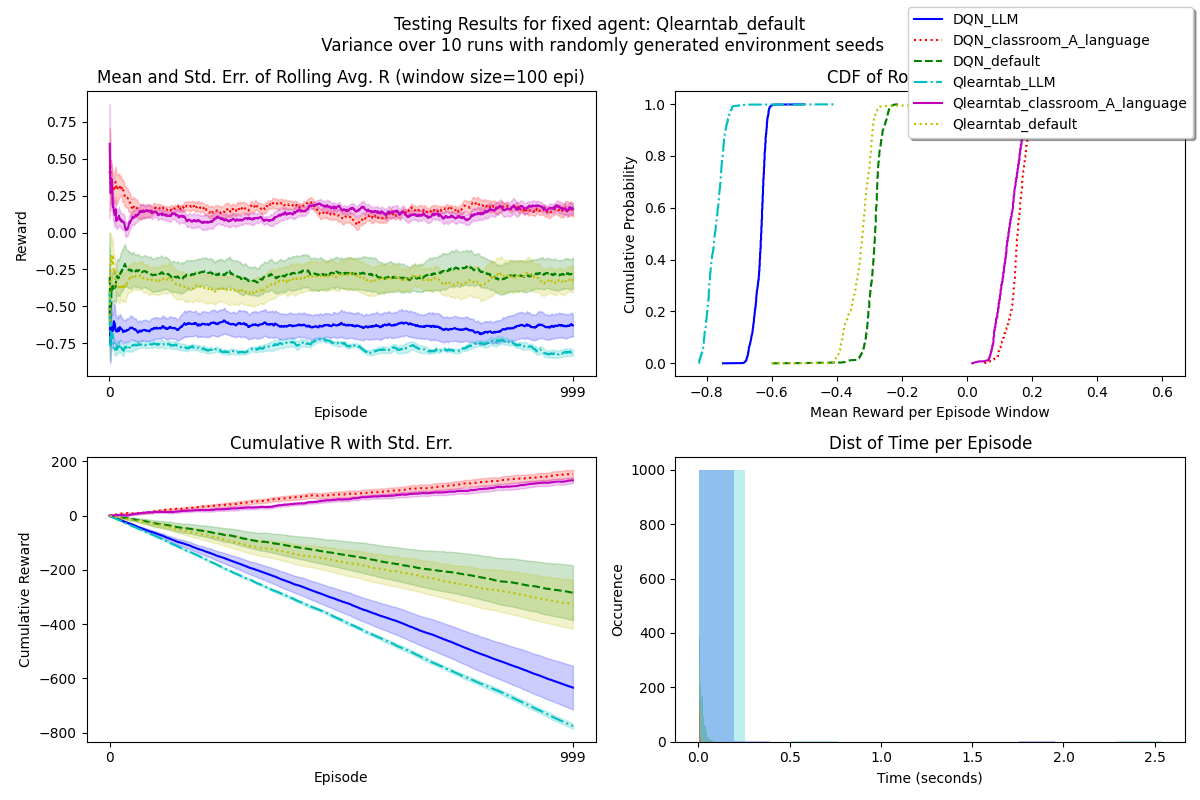}
    \caption{Classroom environment Instruction testing results.}
    \label{fig:results-classroom-test-instr}
\end{figure*}

\begin{figure*}[h!]
    \centering
    \includegraphics[width=1\linewidth]{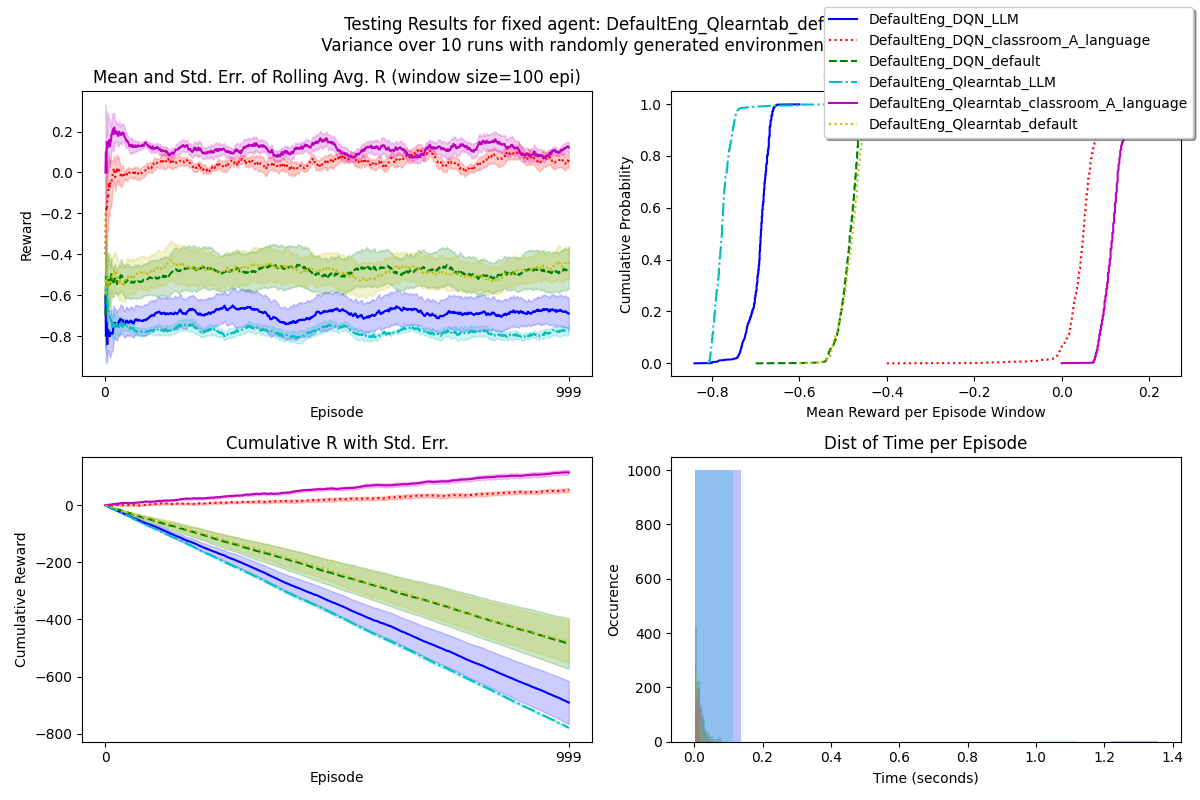}
    \caption{Classroom environment NO instruction training results.}
    \label{fig:results-classroom-test-noinstr}
\end{figure*}

\textbf{Classroom Results}
\begin{itemize}
    \item Fig \ref{fig:results-classroom-train-instr}: Instructions - training
    \item Fig \ref{fig:results-classroom-train-noinstr}: NO instructions - training
    \item Fig \ref{fig:results-classroom-test-instr}: Instructions - testing
    \item Fig \ref{fig:results-classroom-test-noinstr}: NO instructions - testing
\end{itemize}

\begin{figure*}[h!]
    \centering
    \includegraphics[width=1\linewidth]{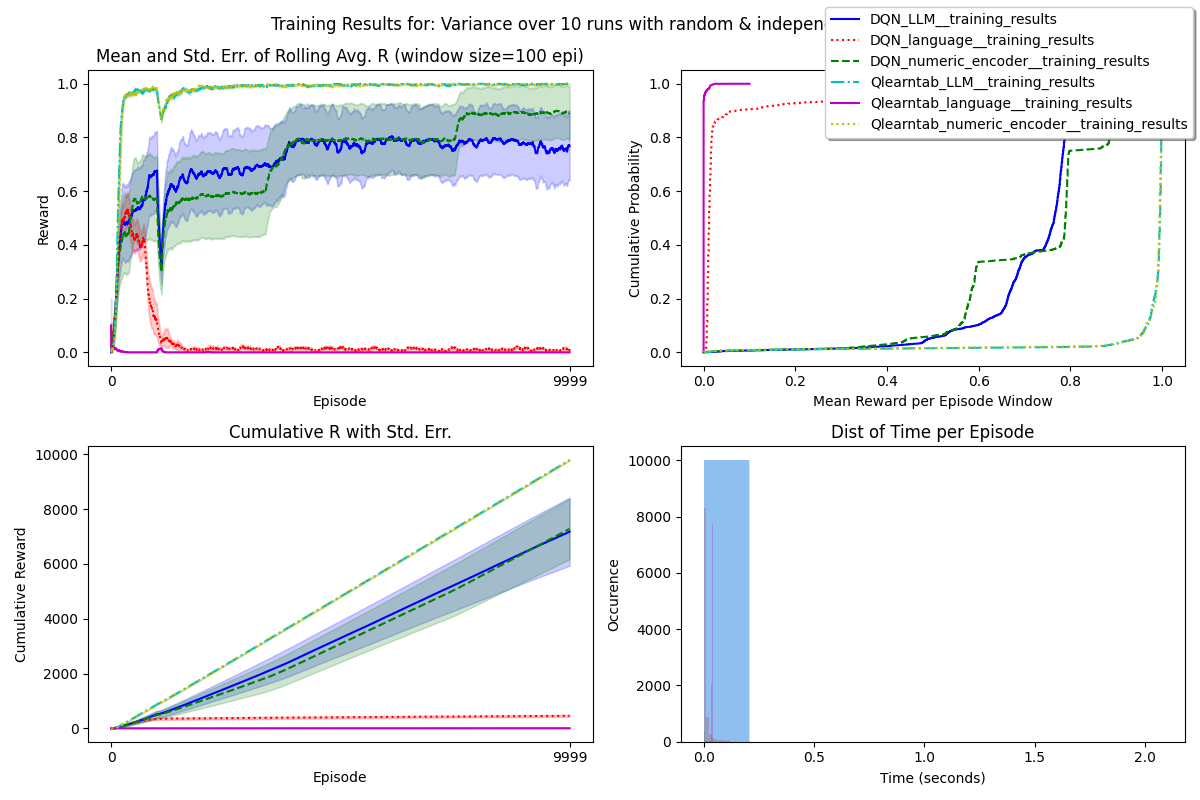}
    \caption{Gym FrozenLake environment Instruction training results.}
    \label{fig:results-GymFrozenLake-train-instr}
\end{figure*}

\begin{figure*}[h!]
    \centering
    \includegraphics[width=1\linewidth]{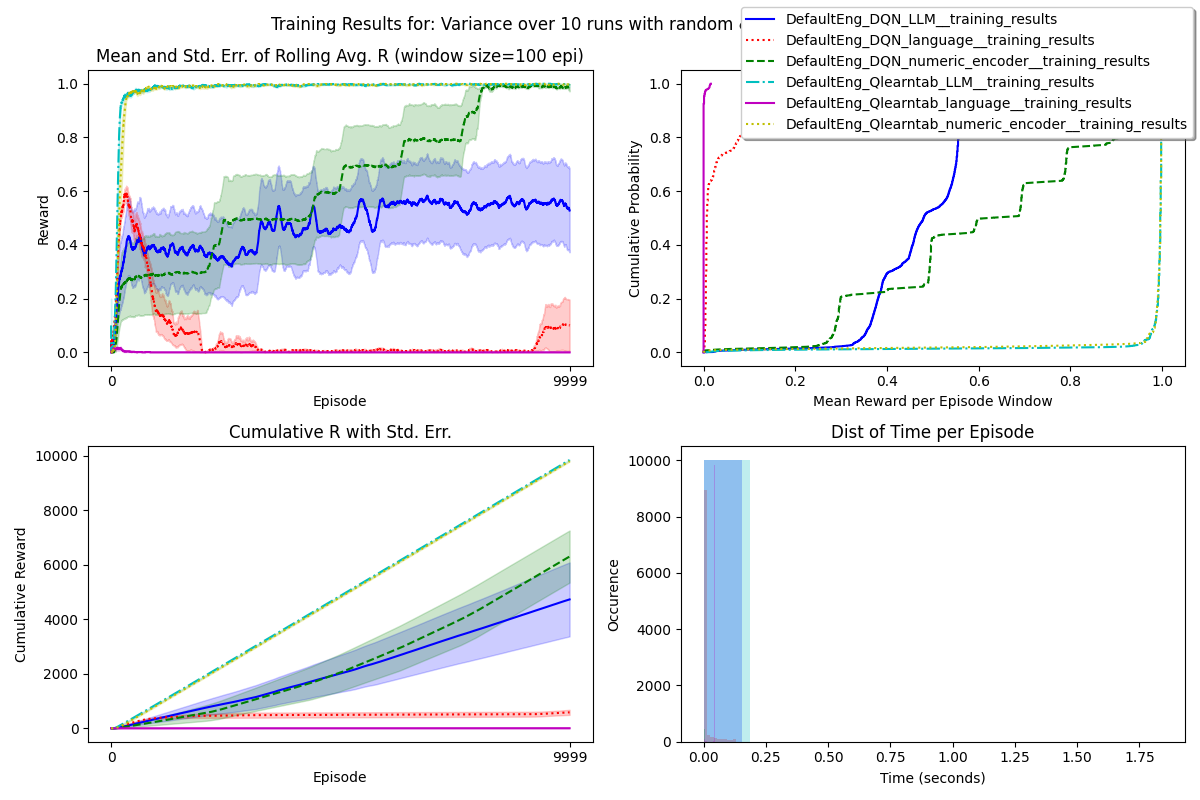}
    \caption{Gym FrozenLake environment NO instruction training results.}
    \label{fig:results-GymFrozenLake-train-noinstr}
\end{figure*}

\begin{figure*}[h!]
    \centering
    \includegraphics[width=1\linewidth]{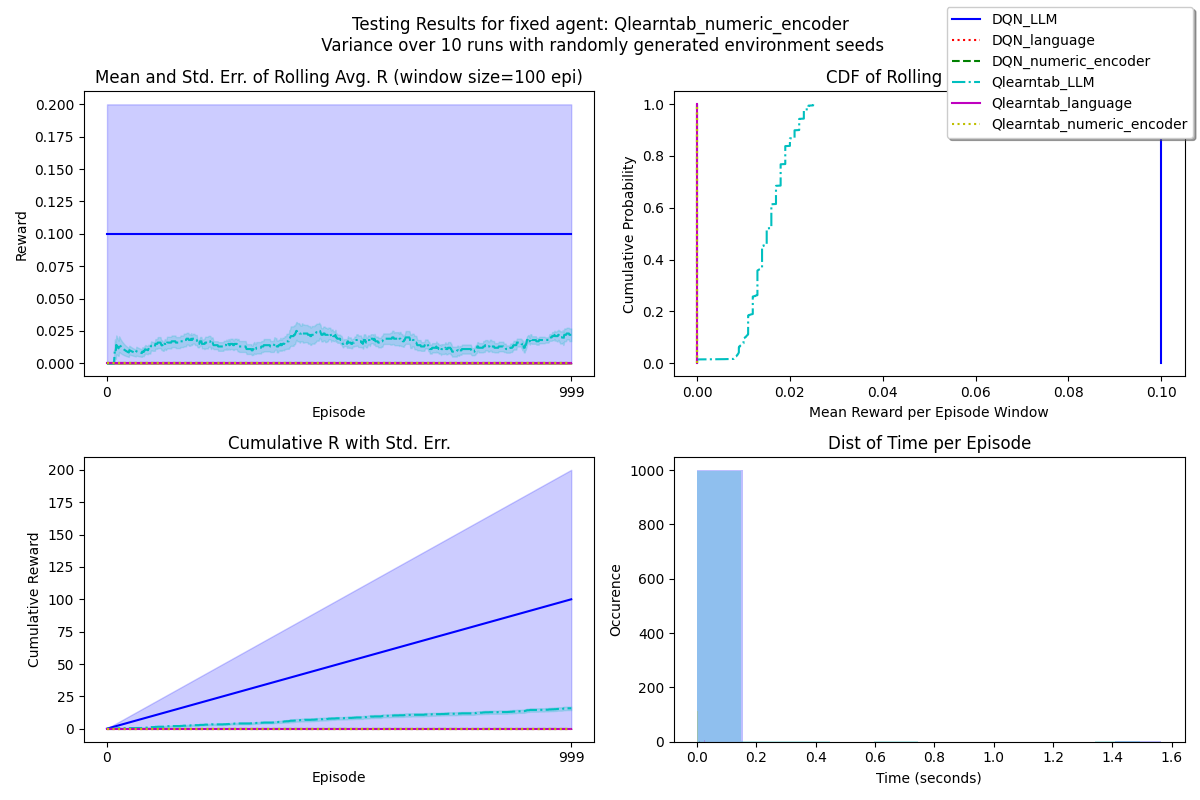}
    \caption{Gym FrozenLake environment Instruction testing results.}
    \label{fig:results-GymFrozenLake-test-instr}
\end{figure*}

\begin{figure*}[h!]
    \centering
    \includegraphics[width=1\linewidth]{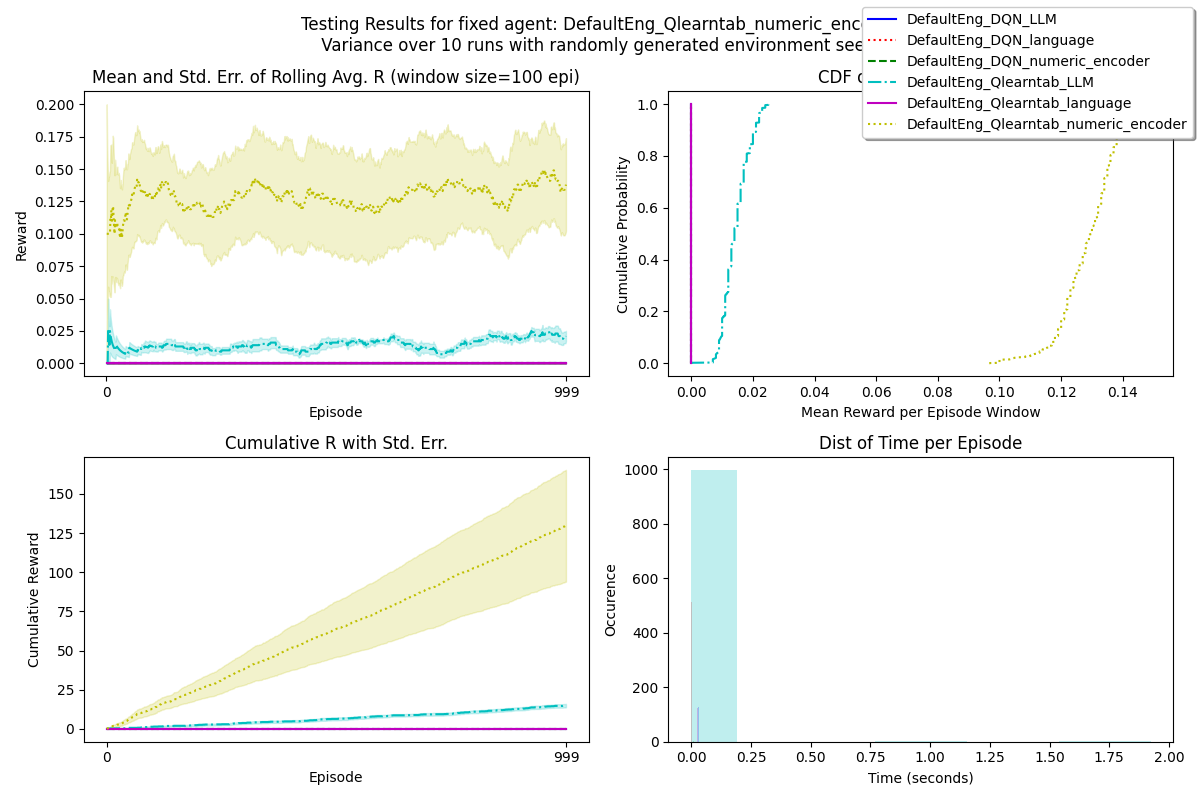}
    \caption{Gym FrozenLake environment NO instruction training results.}
    \label{fig:results-GymFrozenLake-test-noinstr}
\end{figure*}

\textbf{Gym FrozenLake Results}
\begin{itemize}
    \item Fig \ref{fig:results-GymFrozenLake-train-instr}: Instructions - training
    \item Fig \ref{fig:results-GymFrozenLake-train-noinstr}: NO instructions - training
    \item Fig \ref{fig:results-GymFrozenLake-test-instr}: Instructions - testing
    \item Fig \ref{fig:results-GymFrozenLake-test-noinstr}: NO instructions - testing
\end{itemize}

\begin{figure*}[h!]
    \centering
    \includegraphics[width=1\linewidth]{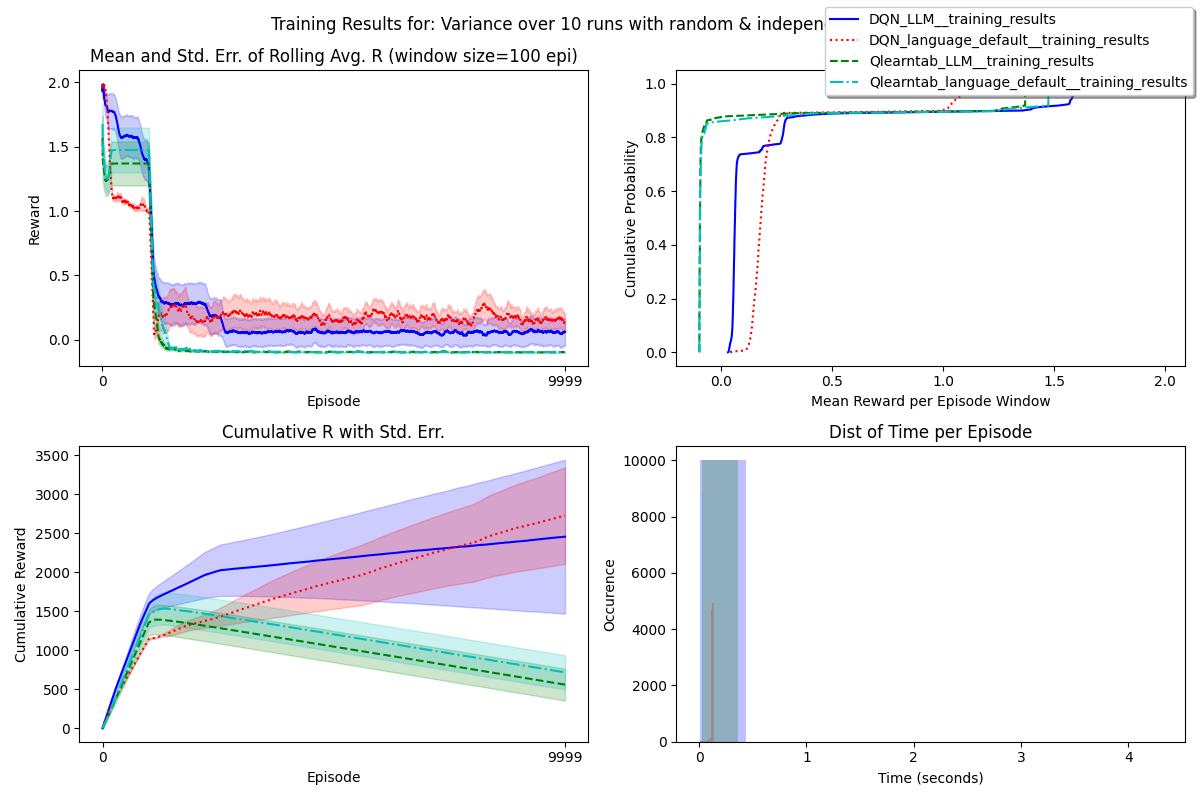}
    \caption{Umaze environment Instruction training results.}
    \label{fig:results-umaze-train-instr}
\end{figure*}

\begin{figure*}[h!]
    \centering
    \includegraphics[width=1\linewidth]{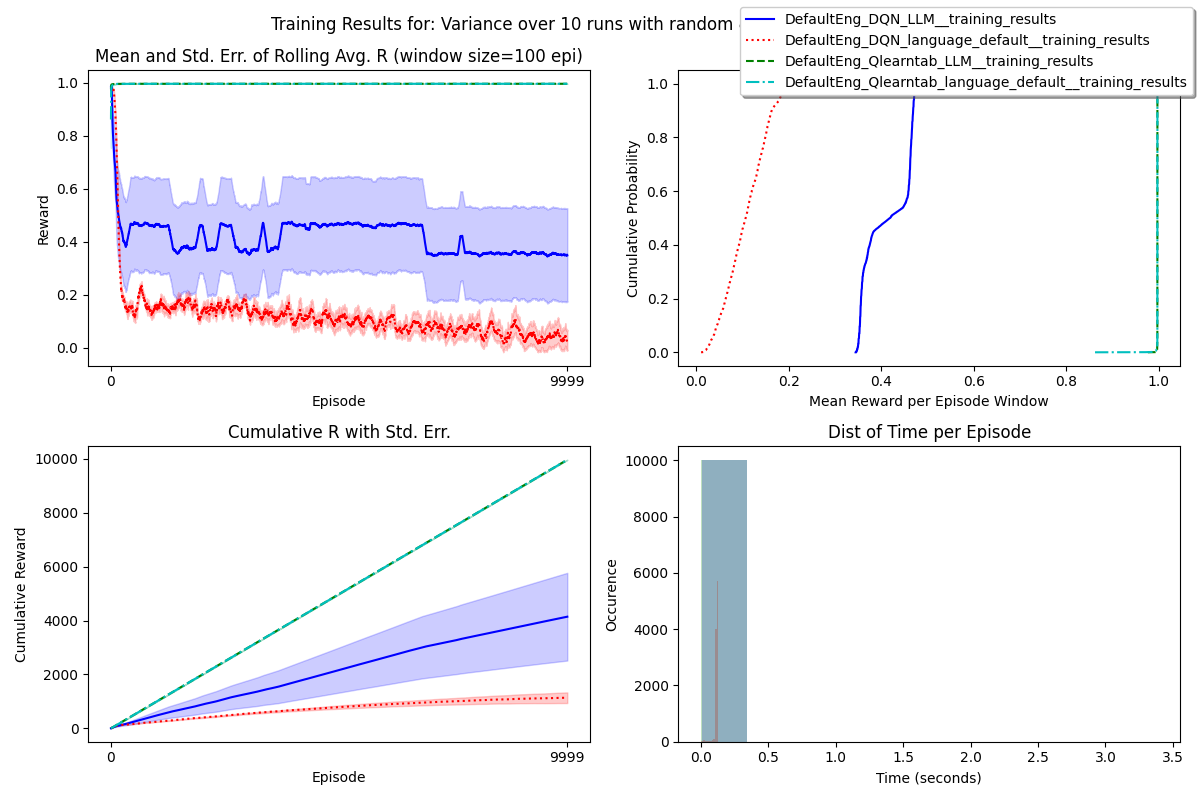}
    \caption{Umaze environment NO instruction training results.}
    \label{fig:results-umaze-train-noinstr}
\end{figure*}

\begin{figure*}[h!]
    \centering
    \includegraphics[width=1\linewidth]{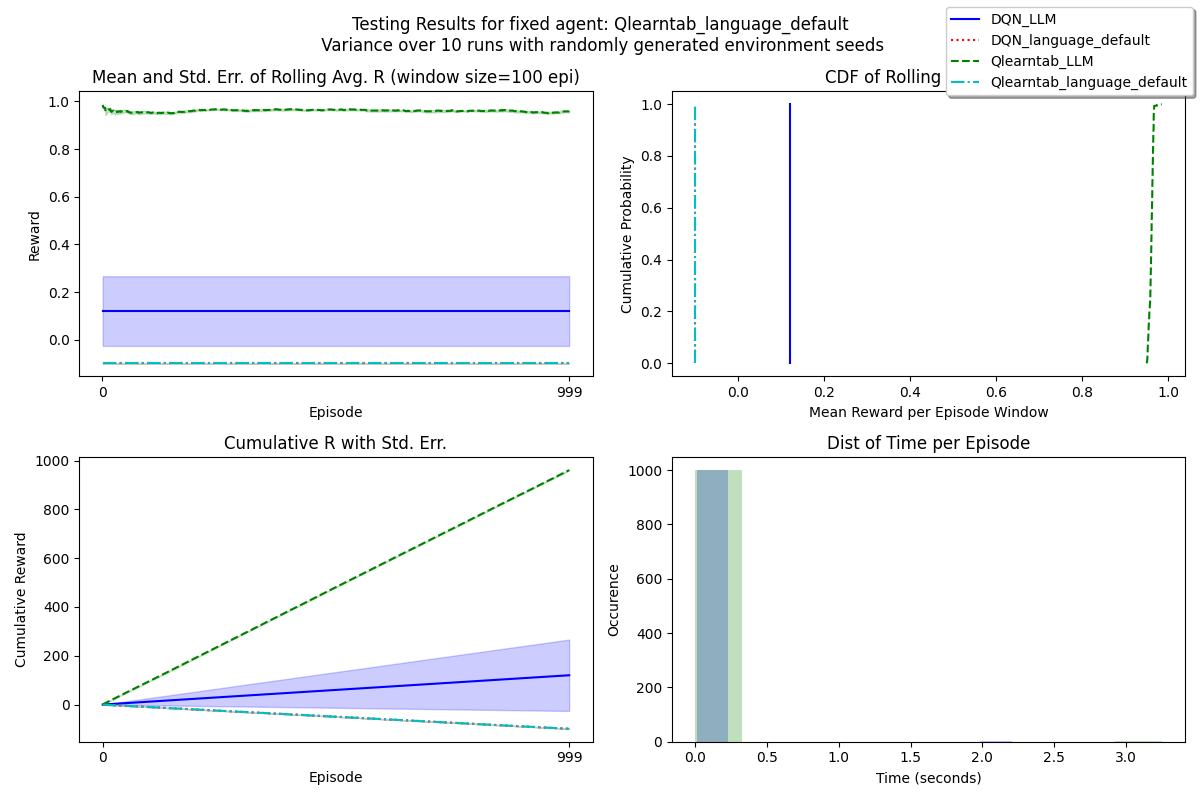}
    \caption{Umaze environment Instruction testing results.}
    \label{fig:results-umaze-test-instr}
\end{figure*}

\begin{figure*}[h!]
    \centering
    \includegraphics[width=1\linewidth]{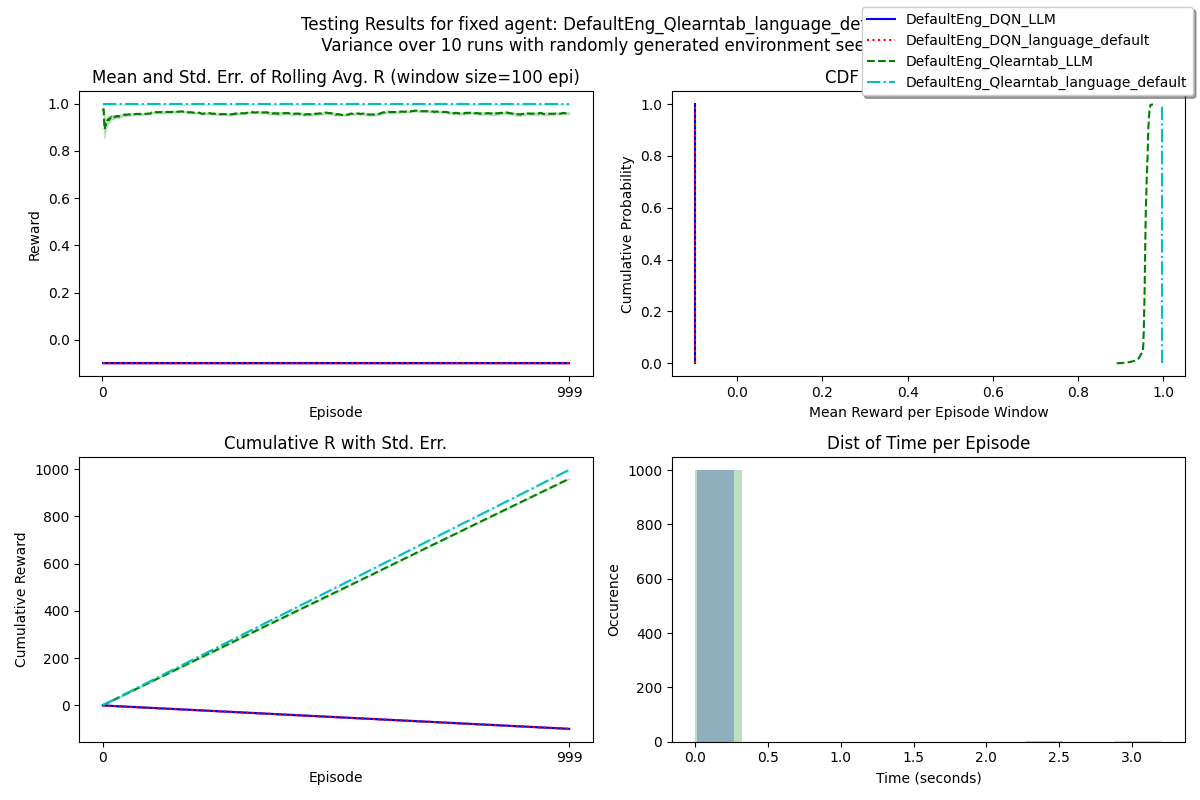}
    \caption{Umaze environment NO instruction training results.}
    \label{fig:results-umaze-test-noinstr}
\end{figure*}

\textbf{Umaze Results}
\begin{itemize}
    \item Fig \ref{fig:results-umaze-train-instr}: Instructions - training
    \item Fig \ref{fig:results-umaze-train-noinstr}: NO instructions - training
    \item Fig \ref{fig:results-umaze-test-instr}: Instructions - testing
    \item Fig \ref{fig:results-umaze-test-noinstr}: NO instructions - testing
\end{itemize}

\begin{figure*}[h!]
    \centering
    \includegraphics[width=1\linewidth]{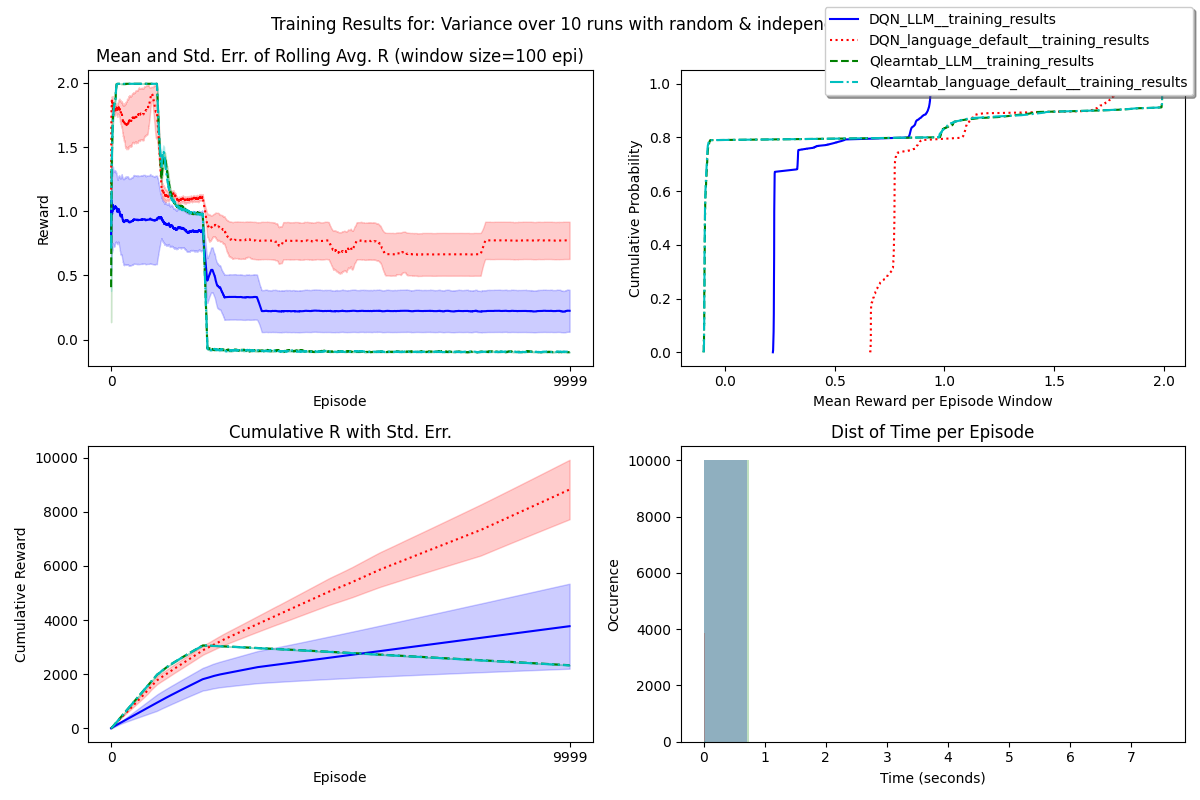}
    \caption{Double-t maze environment Instruction training results.}
    \label{fig:results-double-t-maze-train-instr}
\end{figure*}

\begin{figure*}[h!]
    \centering
    \includegraphics[width=1\linewidth]{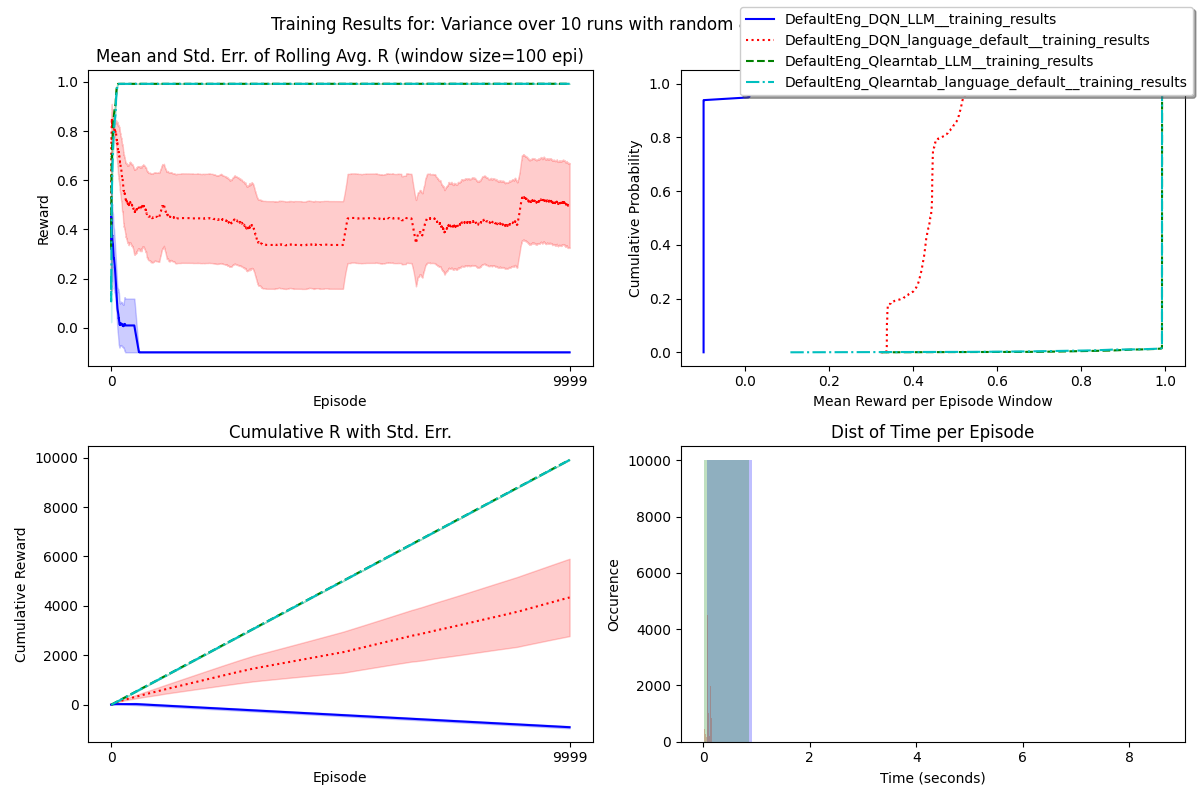}
    \caption{Double-t maze environment NO instruction training results.}
    \label{fig:results-double-t-maze-train-noinstr}
\end{figure*}

\begin{figure*}[h!]
    \centering
    \includegraphics[width=1\linewidth]{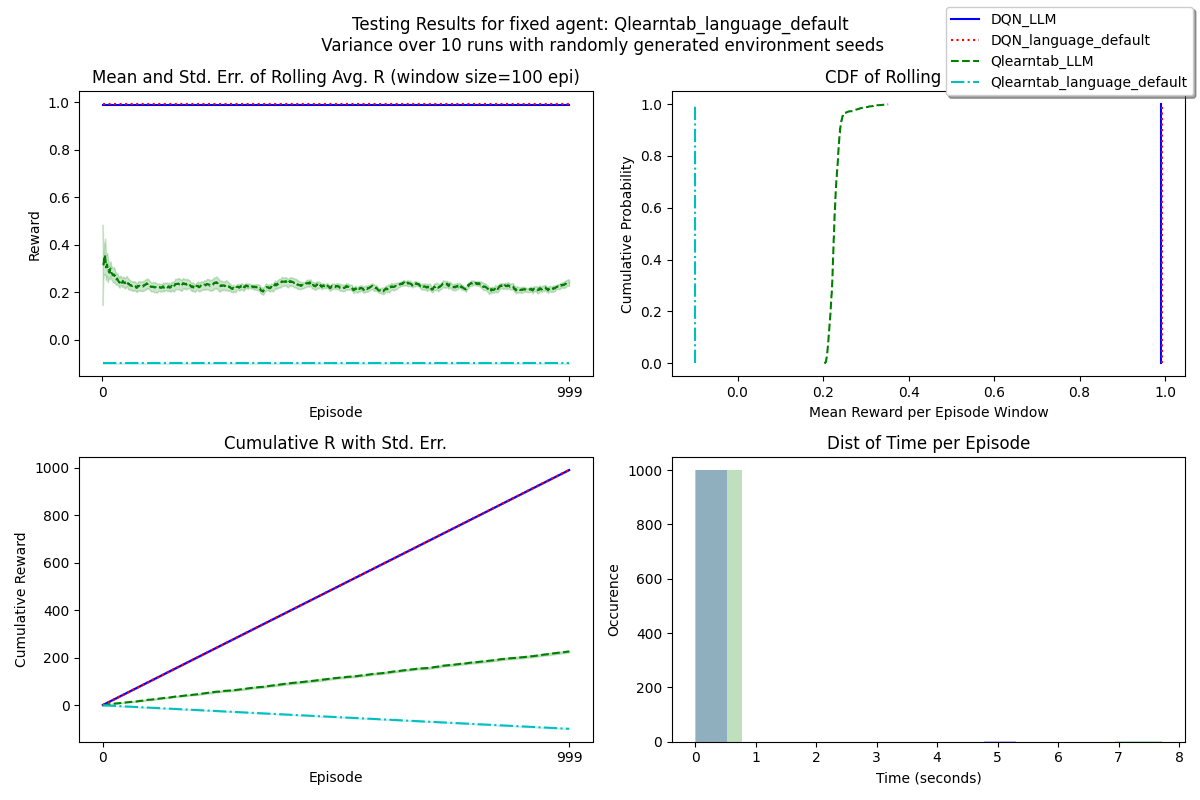}
    \caption{Double-t maze environment Instruction testing results.}
    \label{fig:results-double-t-maze-test-instr}
\end{figure*}

\begin{figure*}[h!]
    \centering
    \includegraphics[width=1\linewidth]{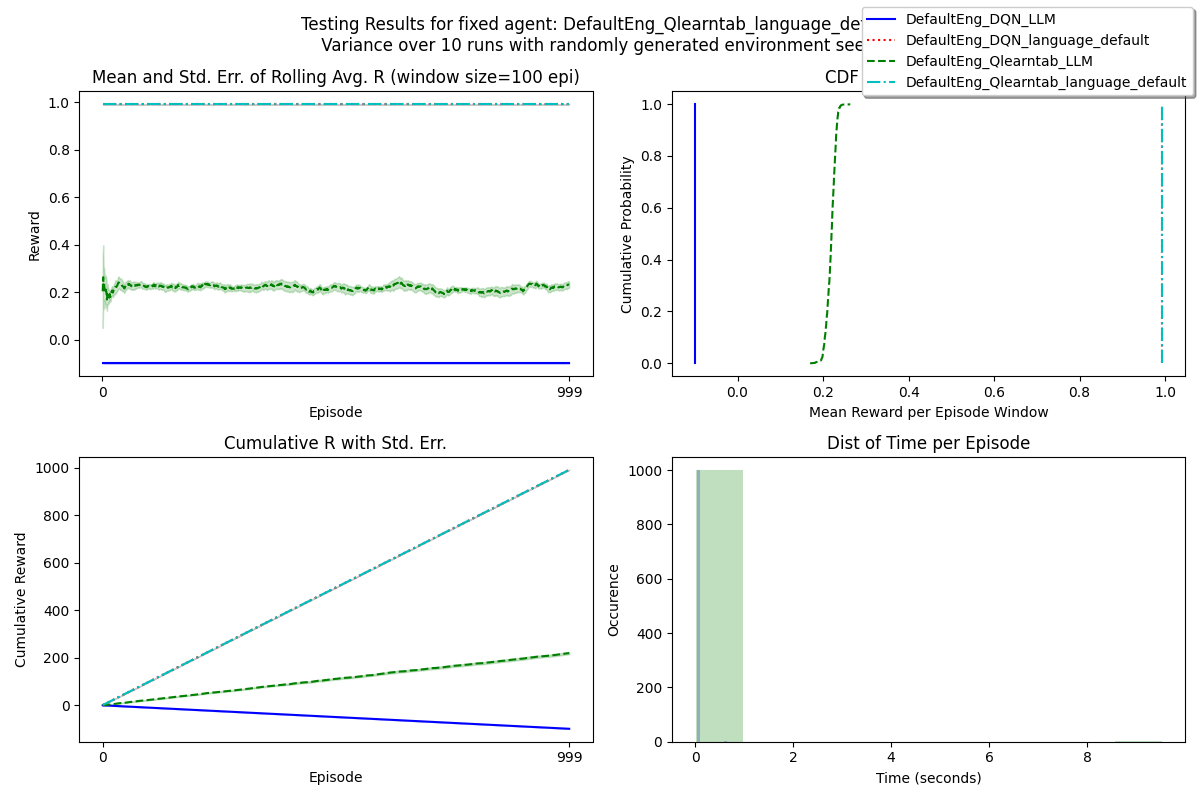}
    \caption{Double-t maze environment NO instruction training results.}
    \label{fig:results-double-t-maze-test-noinstr}
\end{figure*}

\textbf{Double-T Maze Results}
\begin{itemize}
    \item Fig \ref{fig:results-double-t-maze-train-instr}: Instructions - training
    \item Fig \ref{fig:results-double-t-maze-train-noinstr}: NO instructions - training
    \item Fig \ref{fig:results-double-t-maze-test-instr}: Instructions - testing
    \item Fig \ref{fig:results-double-t-maze-test-noinstr}: NO instructions - testing
\end{itemize}

% ---------------------------
% CLASSROOM

% ---------------------------
% GYM FROZEN LAKE

% ---------------------------
% UMAZE

% ---------------------------
% DOUBLE-T-MAZE

\end{document}